\documentclass{article}
\pdfoutput=1  
\usepackage{iclr2027_conference,times}

\usepackage{amsmath,amssymb}
\usepackage{graphicx}
\usepackage{booktabs}
\usepackage{algorithm}
\usepackage{algpseudocode}
\usepackage{float}
\floatstyle{ruled}
\restylefloat{algorithm}
\usepackage{placeins}
\usepackage{afterpage}
\usepackage{xcolor}
\usepackage{colortbl}
\usepackage{tcolorbox}
\definecolor{ourframe}{HTML}{2B5CA8}
\definecolor{ourborder}{HTML}{C6DCF0}
\definecolor{ourback}{HTML}{F1F7FD}
\definecolor{algstream}{HTML}{2B5CA8}
\definecolor{fullcol}{HTML}{F1F7FD} 
\usepackage{url}
\usepackage[colorlinks=true,linkcolor=blue,citecolor=blue,urlcolor=blue]{hyperref}

\newcommand{\method}{\texttt{AttSVD}}
\newcommand{\R}{\mathbb{R}}
\DeclareMathOperator*{\argmin}{arg\,min}
\DeclareMathOperator{\svd}{SVD}

\algrenewcommand\algorithmiccomment[1]{\hfill{\color{gray!55}\textit{#1}}}
\newcommand{\phase}[1]{\Statex\vspace{1.5pt}\hspace*{-1em}{\color{algstream}\textsc{\textbf{#1}}}\vspace{1pt}}

\newcommand{\keybox}[1]{\begin{tcolorbox}[colback=ourback,colframe=ourborder,boxrule=0.8pt,%
  arc=4pt,left=6pt,right=6pt,top=4pt,bottom=4pt,boxsep=2pt,before skip=5pt,after skip=5pt]%
  \small #1\end{tcolorbox}}

\title{\method: Prompt-Adaptive Low-Rank KV Cache Compression via Attention-Guided SVD}

\author{Sara Abdali \quad Jongwoo Ko \quad Pashmina Cameron\\
Microsoft Applied Sciences Group (ASG)\\
\texttt{\{saraabdali, jongwooko, pcameron\}@microsoft.com}}

\begin{document}
\iclrfinalcopy
\maketitle
\lhead{Manuscript under review}\rhead{}

\begin{abstract}
The key-value (KV) cache of autoregressive transformers grows linearly with context length and dominates memory at long context. Most training-free remedies evict low-importance tokens, an irreversible choice along the sequence axis. We instead keep every token and store it more cheaply along the ``\emph{feature}'' axis. We therefore propose \method, a new \emph{interpretable} low-rank compression whose basis is derived from each prompt's own attention geometry: an online, per-prompt truncated SVD that keeps only the directions attention actually reads, cutting persistent per-head KV memory in proportion to the retained rank. We propose two decode-time caching strategies, accumulating and streaming, for short and long generation regimes. Furthermore, we propose two refinements that make compression adaptive. A per-matrix energy rule sizes the logit space and the attention mass independently. An attention-aware basis truncates only in the spaces attention actually reads, preserving both the attention logits and the attention output. The same factors also provide free, per-head interpretability insights into the effective rank and the geometry attention consumes. Across multiple models, on both an agentic benchmark and the full LongBench suite \method\ stays on par with the dense cache while using up to ${\sim}50\%$ of the KV-cache memory.
\end{abstract}

\section{Introduction}
Training-free KV-cache compression methods fall into two families.
\emph{Eviction} methods (e.g.\ H2O~\citep{h2o}, SnapKV~\citep{snapkv}, KeyDiff~\citep{keydiff},
KVzip~\citep{kvzip}) keep a budget of tokens and drop the rest entirely; they differ only in
how they score importance. Eviction is a hard, irreversible choice along the
sequence axis: a wrongly dropped token can never be attended to again.
\emph{Reconstruction} methods (e.g.\ VECTOR~\citep{vector}, Palu~\citep{palu}) instead keep all positions but store them approximately.

We propose \method, which is not only a feature-axis compression method but also introduces two
decode-time caching strategies for different generation regimes. The core rationale builds on the
observed fact that a prompt's keys and values are
effectively low-rank, the keys especially~\citep{eigen,shadowkv}; we extend this low-rank observation
from the $K$ and $V$ spaces to the attention logit space and the
attention mass space. \method\ computes an online
attention-aware factorization in the subspace attention actually reads
(rather than by plain spectral energy), per layer and per key/value head, both at prefill (factoring the prompt bulk)
and during decode (the streaming scenario), where generated tokens are factored once they age out of a short recent-token window so
long generations stay compressed as well (Figure~\ref{fig:method}). Because the basis is the SVD of
the prompt's own activations, it is, by Eckart-Young~\citep{eckartyoung1936}, the best possible rank-$r$ approximation of that
prompt's attention logits and attention mass, obtained with no calibration corpus and no training; no token is discarded,
and every position is still attended to, just from reconstructed rank-$r$ factors.
Our contributions are:
\begin{itemize}
\setlength{\itemsep}{2pt}\setlength{\parskip}{0pt}
\item A per-prompt, calibration-free low-rank KV cache that retains all token positions and is exact at full rank, with a dynamic per-matrix rank rule that sizes keys and values separately so memory adapts to each prompt instead of a fixed schedule;
\item An attention-aware basis that truncates in the subspace attention actually reads, preserving both the attention logits and the attention output;
\item Two decode-time caching strategies that both compress at prefill but differ at decode: an accumulating mode for long-prompt, short-decode scenarios (e.g., no-think mode) that keeps generated tokens exact, and a streaming mode for long-prompt, long-decode scenarios (e.g., think mode) that compresses aged-out tokens into thin blocks for a bounded cache.
\item Interpretability at no extra cost: because the basis is derived from each prompt's own attention geometry, the retained per-matrix ranks and attention-mass profile also serve as per-head diagnostics, exposing each head's effective-rank compressibility, query-subspace focus, and token saliency without calibration.
\end{itemize}

\section{Related Work}\label{sec:related}
Most low-rank KV-cache methods extract the basis ahead of time and reuse it for every prompt.
\emph{Weight decompositions} factor the key/value projection \emph{weights} offline
(ASVD~\citep{asvd}, Palu~\citep{palu}, LoRC~\citep{lorc}); \emph{calibration} methods estimate the
basis from a corpus (Eigen~Attention~\citep{eigen}, ReCalKV~\citep{recalkv}, KQ-SVD~\citep{kqsvd});
and \emph{cross-layer or trained} methods share or learn it
(xKV~\citep{xkv}, MatryoshkaKV~\citep{matryoshkakv}).
OjaKV~\citep{ojakv} is also online and per-context but \emph{approximate} and \emph{variance-greedy}.
It keeps a running incremental-PCA basis seeded from a small calibration set, using attention only
to pick which tokens stay full-rank rather than to shape its basis.

\paragraph{How \method\ differs.}
Prior methods either extract or learn a \emph{single} basis reused for every prompt (the offline
families above), or track an \emph{approximate} running basis seeded from calibration (OjaKV).
\method\ differs on three axes. \emph{(i)~Per-prompt and exact.} It computes the
Eckart-Young-optimal truncated SVD of \emph{this} prompt's keys and values at prefill, with no
calibration corpus, no training, and no running-basis bookkeeping; the basis is then frozen during
generation (unlike OjaKV's seeded, incremental estimate). \emph{(ii)~Attention-guided, not
variance-greedy.} Rather than keeping the directions of largest $K,V$ variance, it truncates in the
space attention actually consumes; weighting keys by the query covariance $C_Q{=}Q^\top Q$ (which
preserves the logits $QK^\top$) and values by per-token attention mass (which preserves the output
$PV$), so the retained subspace is \emph{functional} rather than merely descriptive. This is a
distinct \emph{form} of attention-awareness from KQ-SVD: rather than low-rank-approximating the
\emph{products} $KQ^\top$ and $VW_O$ learned offline from calibration, \method\ keeps the ordinary
per-matrix factors and merely reweights their SVD metric, so one calibration-free rule aligns
\emph{both} keys (via $C_Q$, preserving the scores) and values (via attention mass, preserving the
output) with \emph{this} prompt's own attention. \emph{(iii)~Dynamic
per-matrix rank.} A spectral-energy rule factors each matrix independently
from the data so that memory adapts to each prompt instead of following a fixed schedule.
For a comprehensive related work discussion, see
Appendix~\ref{app:related}.

\section{Method}\label{sec:method}

\keybox{\textcolor{ourframe}{\textbf{Motivation:}} Existing KV-cache methods save memory by \emph{evicting} tokens,
permanently discarding their keys and values. We instead retain \emph{every} token in a cheaper
form, re-expanded transiently at
inference time, with no eviction or training. Unlike other low-rank reconstruction methods, we aim at a low-rank compression that
is \emph{per-matrix adaptive}, \emph{calibration-free}, \emph{prompt-adaptive}, robust across \emph{prefill \& decode
length scenarios}, and \emph{attention-importance aware}.}

\subsection{Proposed \method}
\paragraph{Setup and notation.}
Let $K, V \in \mathbb{R}^{S \times d}$ denote the
keys and values for a prompt of length $S$, where $d$ is the head dimension.
Standard attention computes $P = \mathrm{softmax}\!\left(QK^\top/\sqrt{d} + M\right)$
and outputs $PV$, where $M$ is the causal mask and $Q\in\R^{S\times d}$ the queries.
A single head stores $2Sd$ entries ($K$ and $V$), for a total KV-cache footprint of $2SdLH$
across $L$ layers and $H$ key/value heads. In case of
grouped-query attention (GQA), $H$ is the number of \emph{key/value} heads, typically a
small multiple fewer than the number of query heads $H_q$; this matters for the attention-aware
variant, where a key head is shared by several query heads.
Throughout the paper, $K$ denotes the \emph{post-RoPE} keys as they sit in the cache,\footnote{Rotary embeddings
rotate $Q$ and $K$ but never $V$.} so our attention-weighted factorization operates on the rotated
keys; this differs from
ShadowKV~\citep{shadowkv}, which compresses the \emph{pre-RoPE} keys because the position-dependent
rotation spreads the keys across more singular directions, so the pre-RoPE keys are lower-rank and
compress better, at the price of storing and replaying the rotation at reconstruction. This
pre-RoPE trick does not apply to us, because we never run an SVD on raw $K$ or $V$; we factor in the
attention-logit and attention-mass geometry instead. The quantity our key basis must preserve is the
logit inner product $QK^\top$, which is exact only when the keys carry the \emph{same} rotation as
the queries, so operating post-RoPE is required and a pre-RoPE key basis would be misaligned with the
post-RoPE query metric that our attention-aware truncation uses. The modestly higher $r_{\widetilde K}$ this positional spreading induces is
in turn absorbed by the per-prompt adaptive rank.

\paragraph{Standard SVD on weighted inputs.} \method\ introduces no new decomposition. It runs the
\emph{same} standard truncated SVD as SVD on raw KV, only on attention-weighted $K$ and $V$ rather
than on $K$ and $V$ directly, so it is standard SVD on weighted inputs, not a separate weighted-SVD
routine.

\begin{figure*}[t]
\centering
\includegraphics[width=\linewidth]{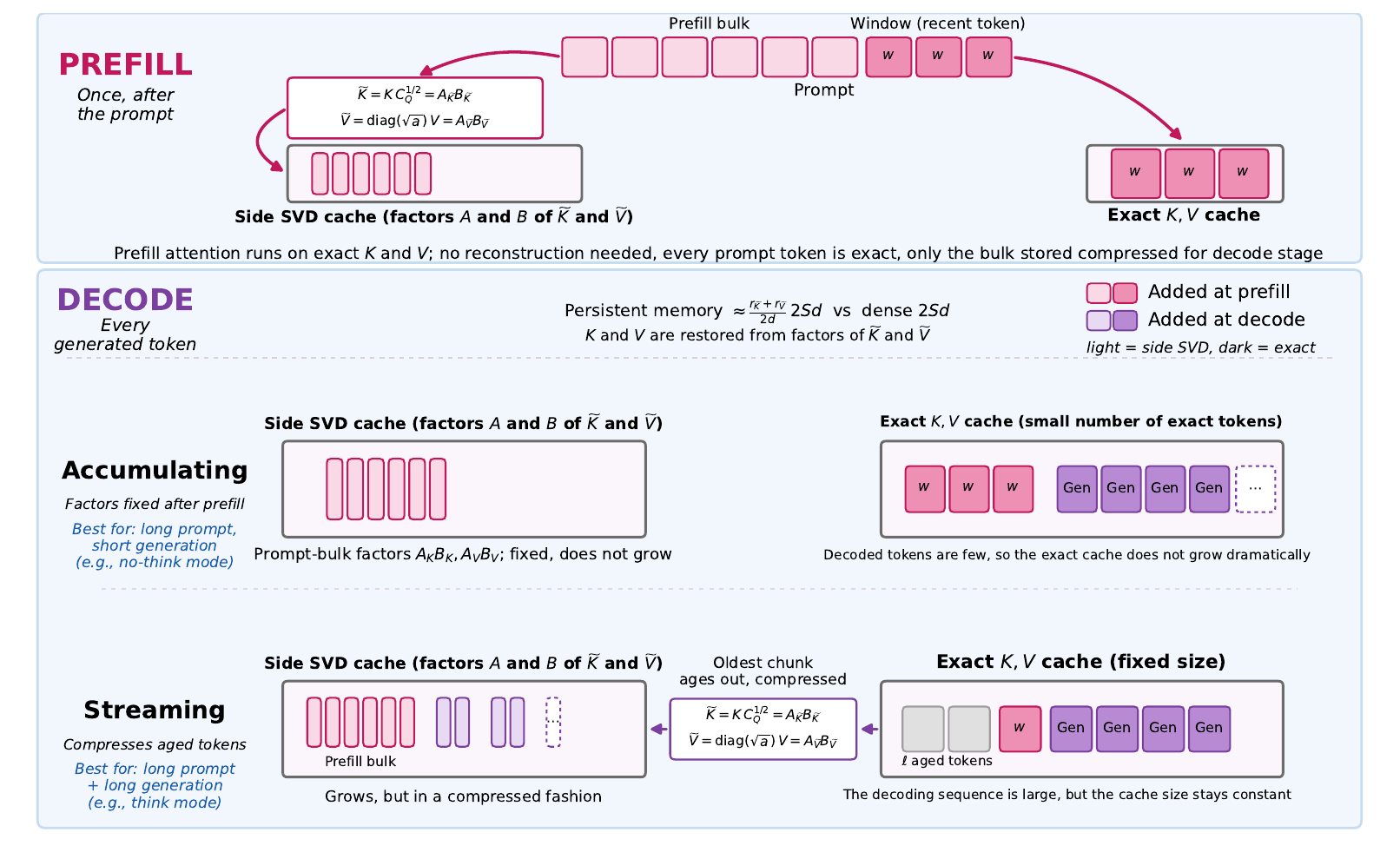}
\caption{\method\ keeps a thin, compressed side cache of attention-aware factors $A,B$ plus a small
exact cache of recent tokens, cutting persistent memory to a fraction of dense while retaining all
token positions. At decode, \emph{accumulating} keeps every token exact (small decoded tokens); \emph{streaming}
compresses tokens leaving a sliding window, keeping the exact buffer bounded.}
\label{fig:method}
\end{figure*}

\subsection{Per-prompt and attention-aware low-rank compression}
\label{sec:attn}
In general, for an $S\times d$ matrix $Z$ with $S>d$ the (full) singular value decomposition (SVD) is
$Z=U\Sigma \mathcal V^\top$, where \emph{at full rank} $U\in\R^{S\times d}$,
$\Sigma=\mathrm{diag}(\sigma_1\ge\cdots\ge\sigma_d\ge0)\in\R^{d\times d}$, and
$\mathcal V\in\R^{d\times d}$ (so $Z$ has at most $d$ nonzero singular values). The rank-$r$
truncation ($r\le d$)
\begin{equation}
Z_r \;=\; U_{:,1:r}\,\Sigma_{1:r}\,\mathcal V_{:,1:r}^\top
\;=\; \argmin_{\mathrm{rank}(X)\le r}\;\lVert Z-X\rVert_F
\label{eq:ey}
\end{equation}
is, by the Eckart-Young theorem, the \emph{optimal} rank-$r$ approximation of $Z$, with
error $\lVert Z-Z_r\rVert_F=\big(\sum_{i>r}\sigma_i^2\big)^{1/2}$, where we store the \emph{factors}
\begin{equation}
A \;=\; U_{:,1:r}\,\Sigma_{1:r}\in\R^{S\times r},
\qquad
B \;=\; \mathcal V_{:,1:r}^\top\in\R^{r\times d},
\qquad
\widetilde Z \;=\; AB .
\label{eq:factors}
\end{equation}
\method\ is attention-aware, i.e.\ the SVD is taken in the score and output spaces: it preserves the
attention logits $QK^\top$ by weighting the keys in the query geometry $C_Q=Q^\top Q$, and the output
$O=PV$ by weighting the values by their per-key
attention mass, changing only the basis, not the factor shapes.
SVD on raw $K$ and $V$ instead minimizes $\lVert K-K_r\rVert_F$ and $\lVert V-V_r\rVert_F$, so it can spend rank on directions attention never queries; we truncate in the spaces attention uses, keeping the decomposed storage of Eq.~\eqref{eq:factors} unchanged. \emph{Either way the decomposition is the same standard
truncated SVD; only its input differs, raw versus attention-weighted $K$ and $V$.}

\paragraph{Weighted keys via the query geometry.} Replacing $K$ by $K_r$ perturbs the logits by
$Q(K-K_r)^\top/\sqrt d$, so the relevant objective is
\begin{equation}
K_r^\star=\argmin_{\mathrm{rank}(X)\le r}\big\lVert Q(K-X)^\top\big\rVert_F
=\argmin_{\mathrm{rank}(X)\le r}\big\lVert (K-X)\,C_Q^{1/2}\big\rVert_F,
\qquad C_Q=Q^\top Q,
\label{eq:kqmetric}
\end{equation}
the optimal rank-$r$ approximation of $K$ in the query-covariance metric $C_Q$ (the two norms are
equal since both are $\mathrm{tr}\!\big((K-X)\,C_Q\,(K-X)^\top\big)$). By the change of variables
$\widetilde K=K\,C_Q^{1/2}$ the generalized Eckart-Young solution is the ordinary truncated SVD of
$\widetilde K$ mapped back: factor $\widetilde K=U\Sigma\mathcal V^\top$ and store
$A_{\widetilde K}=U_{:,1:r}\Sigma_{1:r}$ and $B_{\widetilde K}=\mathcal V_{:,1:r}^\top C_Q^{-1/2}$, so that $A_{\widetilde K}B_{\widetilde K}=K_r^\star$
in the original coordinates. Only the basis changes; the cache layout and decode reconstruction
are identical.

\paragraph{Weighted values via the attention mass.} Replacing $V$ by $V_r$ perturbs the output by $P(V-V_r)$. The
exact minimizer of $\lVert P(V-V_r)\rVert_F$ needs the $S\times S$ Gram $P^\top P$; we use the
diagonal (per-key mass) approximation $a_i=\sum_q P_{qi}$, the total attention received by key $i$,
giving
\begin{equation}
V_r^\star=\argmin_{\mathrm{rank}(X)\le r}\sum_{i=1}^{S} a_i\,\lVert V_i-X_i\rVert_2^2
=\argmin_{\mathrm{rank}(X)\le r}\big\lVert \mathrm{diag}(\sqrt a)\,(V-X)\big\rVert_F,
\label{eq:vmass}
\end{equation}
solved by a change of variables: form the row-weighted values
$\widetilde V=\mathrm{diag}(\sqrt a)\,V=U\Sigma\mathcal V^\top$, whose best rank-$r$ fit is the standard
truncated SVD $\widetilde V_r=U_{:,1:r}\Sigma_{1:r}\mathcal V_{:,1:r}^\top$ (Eq.~\ref{eq:ey}); un-weighting
its rows by $\mathrm{diag}(1/\sqrt a)$ recovers the store
$A_{\widetilde V}=\mathrm{diag}(1/\sqrt a)\,U_{:,1:r}\Sigma_{1:r}$ and $B_{\widetilde V}=\mathcal V_{:,1:r}^\top$, so that
$A_{\widetilde V}B_{\widetilde V}=\mathrm{diag}(1/\sqrt a)\,\widetilde V_r=V_r^\star$. In both cases the kept directions are
exactly the ones attention reads; full proofs of both truncations are given in
Appendix~\ref{app:kqproof}.

\paragraph{Obtaining $Q$ and $a$.} Both are available at prefill: we capture the \emph{post-RoPE}
queries with lightweight forward hooks (a short observation window per layer) and read the value
mass $a$ from one attention-shaped pass, with GQA expansion and small numerical guards for the
un-weighting. This composes with any rank rule of Section~\ref{sec:adaptive}: the weighting
sets the basis, the rule sets the rank.

\subsection{Two KV-cache compression regimes tailored to prefill and decode length}
\label{sec:static}
\method\ always compresses the prompt at \emph{prefill}; the two regimes differ only in how the
cache is maintained during \emph{decode}. We first describe the shared prefill phase, then the two
decode variants: an \emph{accumulating} and a \emph{streaming}~cache.

\paragraph{Prefill.} Given the prompt's $\widetilde K,\widetilde V$ for a layer, split off a small \emph{recent
window} of the newest $w$ tokens, kept exact; call the remaining bulk
$K^{(b)},V^{(b)}\in\R^{(S-w)\times d}$. AttentionSVD each via Eq.~\eqref{eq:factors} into $(A_{\widetilde K},B_{\widetilde K})$
and $(A_{\widetilde V},B_{\widetilde V})$ and store \emph{only the factors} in a compressed side cache, writing just the exact
tail to the exact cache. The prefill forward itself is served with the \emph{exact}
$K,V$ that are already in hand; reconstruction at prefill would only inject a lossy step we do
not need, so prefill output is bit-identical to the dense cache. The compressed factors are a lossy
approximation, consulted only when those tokens are later reconstructed at decode.

\paragraph{Exact recent-window tail.}
\label{sec:tail}
The newest $w$ tokens are stored exactly rather than decomposed. They carry
the highest-impact local context, and a handful of tokens cannot be usefully compressed by
a rank-$r$ SVD (and a single decode token is appended one at a time, where factoring is
counter-productive). Setting $w=0$ recovers a fully decomposed prefill.

\paragraph{Decode.} At each generation step the decomposed bulk is re-materialized transiently
as $\widehat K^{(b)}=A_{\widetilde K}B_{\widetilde K}\in\R^{(S-w)\times d}$ and $\widehat V^{(b)}=A_{\widetilde V}B_{\widetilde V}$
(the stored $B$ factors carry the reverse map, so these products approximate the \emph{original}
$K^{(b)},V^{(b)}$, not $\widetilde K,\widetilde V$), concatenated with the exact tail and consumed by
attention, which forms $P$ from $QK^\top$ and $O=PV$ on the recovered $K,V$; these tensors exist for one layer at a
time and are freed on return, so the transient peak is $\mathcal O(Sd)$, never the whole cache. Only
the low-rank factors persist across steps. The two strategies differ in decode-time caching:

\paragraph{Decode strategy 1: accumulating cache.} The prefill factors are fixed and each new token
is appended \emph{exactly} to the cache, never compressed; no decode-time SVD is performed and the
decode-factor side cache stays empty, so the cache grows with generation length. This is best when the
prompt is long but generation is short (e.g.\ multiple-choice tasks or \emph{no-think} mode), where
the exact generated tokens stay a small fraction of the cache.

\paragraph{Decode strategy 2: streaming cache (sliding window).}\label{sec:streaming} Streaming
\method\ bounds the decode-time footprint by compressing \emph{during} decode
(Figure~\ref{fig:method}, bottom), holding the cache at a \emph{fixed} size regardless of output
length, which suits long \emph{think}-mode generations. It maintains a \emph{sliding} exact window of
the most recent $w$ tokens: each generated token is appended to the window, and once the window
overflows by a full chunk of $\ell$ tokens (the \texttt{streaming} branch of
Algorithm~\ref{alg:main}, App.~\ref{app:algorithms}), the oldest $\ell$ tokens (still held \emph{exactly}) are factored with the
same decomposition into a per-layer list of frozen \emph{decode blocks}, then dropped from the
window. Decode attention runs on the full sequence, formed by concatenation ($\oplus$) of prefill bulk
$\oplus$ decode blocks $\oplus$ exact window in position order. Two properties keep it sound: aged-out tokens are factored from their \emph{exact}
values (never by compressing the already-lossy prefill bulk), so error does not compound across the
generation, and each block is frozen, so a token is compressed at most once. The scheme composes with
any rank rule of Section~\ref{sec:adaptive} and the attention-aware basis of Section~\ref{sec:attn}.

\subsection{Choosing the rank: compression-ratio or per-matrix energy}
\label{sec:adaptive}
\method\ offers two strategies for setting the rank, each suited to a different use case: a
\emph{(1) fixed compression ratio} $c$ when a specific memory budget is required, and a
\emph{(2) per-matrix energy rule} when factor denoising and minimizing reconstruction error matter
more than hitting a target budget.

\paragraph{Compression-ratio rank (unified).} If an absolute rank $r$ or a compression ratio $c$
is supplied, both $K$ and $V$ use the same $r$ (with $r=\mathrm{round}((1-c)d)$ for a ratio),
recovering the basic decomposition of Eq.~\eqref{eq:factors}. This \emph{uniform} setting trades away
per-matrix adaptivity for an exact, predictable budget; note that $c$ selects a \emph{rank} (a
fraction of the $d$ feature directions), not a fraction of \emph{tokens}, so all $S$ tokens are
retained. This is the operating point used
for the matched-memory settings in Section~\ref{sec:ablation} (e.g.\ $c{=}0.5\Rightarrow r{=}64$).
The two ranks may also be set individually as explicit $r_{\widetilde K}$ and $r_{\widetilde V}$, overriding the shared
value; this exposes the per-matrix budget as a direct hyperparameter and enables controlled study
of the asymmetric $r_{\widetilde K}\!\ll\!r_{\widetilde V}$ regime.

\paragraph{Per-matrix energy rank (dynamic).} Keys and values are not equally compressible.
Building on the pre-RoPE-key observation of ShadowKV~\citep{shadowkv} that keys and values differ in
their intrinsic compression capacity, the attention-weighted matrices we factor,
$\widetilde K$ and $\widetilde V$, likewise likely differ, so a single shared rank over-spends on $\widetilde K$ and
under-serves $\widetilde V$ (or vice versa) and the two may need different ranks. Here
\method\ lets each matrix choose its own rank from its own spectrum. Ordering a matrix $M$'s
singular values $\sigma_1\!\ge\!\sigma_2\!\ge\!\cdots$, we keep the fewest leading directions that
retain a fraction $\tau$ of its energy $\lVert M\rVert_F^2{=}\sum_i\sigma_i^2$,
\begin{equation}
r_M(\tau)\;=\;\min\Big\{\,r:\ \textstyle\sum_{i\le r}\sigma_i^2(M)\ \ge\ \tau\sum_i\sigma_i^2(M)\,\Big\},
\qquad M\in\{\widetilde K,\widetilde V\},
\label{eq:energyrank}
\end{equation}
the \emph{effective rank} of $M$ at tolerance $\tau$~\citep{roy} (default $0.99$): a fast-decaying
spectrum reaches $\tau$ in a few directions, a flat one needs many. Applying the \emph{same} $\tau$
to both matrices yields per-matrix ranks $r_{\widetilde K}$ and $r_{\widetilde V}$ that adapt to each
prompt's own spectrum rather than being hand-set, and may differ from one another. Equation~\eqref{eq:energyrank} gives one rank per $(\text{batch},\text{head})$; since the
cache stores a single rank per matrix, these are reduced by a configurable aggregator:
\texttt{max} (the default; the largest, so \emph{every} head retains at least $\tau$),
\texttt{mean}, or a percentile \texttt{p}$N$ (allowing a few low-energy heads to fall below $\tau$
to save memory). An optional \texttt{rank\_max} caps the result; passing an explicit $\tau$ selects
this dynamic mode even when a rank or ratio is otherwise set.

\subsection{Memory cost}
\label{sec:memory}
The saving is explicit. A dense head stores
$K,V\in\R^{S\times d}$, i.e.\ $2Sd$ entries; \method\ instead keeps the factors that reconstruct
them. For the prompt bulk it stores the factors
$A_{\widetilde K}\in\R^{(S-w)\times r_{\widetilde K}},B_{\widetilde K}\in\R^{r_{\widetilde K}\times d}$ for the keys
and $A_{\widetilde V}\in\R^{(S-w)\times r_{\widetilde V}},B_{\widetilde V}\in\R^{r_{\widetilde V}\times d}$ for the values.
A rank-$r$ factorization of an $(S{-}w)\times d$ matrix costs $r\,(S{-}w{+}d)$ entries, so the two
factor pairs together use $(r_{\widetilde K}{+}r_{\widetilde V})(S{-}w{+}d)$ entries, plus a short exact tail
of the $w$ most recent tokens ($2wd$ entries). The tail is $\mathcal{O}(wd)$ with $w$ fixed and
$w\!\ll\!S$, so it is negligible against the bulk; dropping it and using $S{-}w\!\approx\!S$ leaves a
persistent factor cost of $(r_{\widetilde K}{+}r_{\widetilde V})(S{+}d)$. Dividing by the dense $2Sd$ and
expanding,
\begin{equation}
\frac{(r_{\widetilde K}{+}r_{\widetilde V})(S{+}d)}{2Sd}
\;=\; \frac{r_{\widetilde K}{+}r_{\widetilde V}}{2d}\Big(1+\frac{d}{S}\Big)
\;=\; \frac{r_{\widetilde K}{+}r_{\widetilde V}}{2d} + \frac{r_{\widetilde K}{+}r_{\widetilde V}}{2S}
\;\xrightarrow[S\gg d]{}\; \frac{r_{\widetilde K}{+}r_{\widetilde V}}{2d}.
\label{eq:ratio}
\end{equation}
The residual $(r_{\widetilde K}{+}r_{\widetilde V})/(2S)$ term vanishes once the context is long ($S\gg d$),
leaving the compression ratio $(r_{\widetilde K}{+}r_{\widetilde V})/(2d)$. The asymmetric
regime $r_{\widetilde K}\!\ll\!r_{\widetilde V}$ shrinks this further, while a shared rank
$r_{\widetilde K}{=}r_{\widetilde V}{=}r$ reduces it to simply $r/d$.
The recent window adds a further $\mathcal{O}(w/S)$ term, and as context grows the measured
ratio tightens to $r/d$. At decode,
the \emph{accumulating} cache keeps every generated token exact, so its exact part is the recent tail
plus all $g$ generated tokens, $\mathcal{O}\big((w{+}g)d\big)$; the \emph{streaming} cache instead
bounds it, oscillating in $[w,\,w{+}\ell]$ independent of $g$, while the compressed part grows by one
$\mathcal{O}(\ell r+r d)$ block per $\ell$ tokens ($\approx r/d$ of dense per token once $\ell\gg d$,
the same asymptotic ratio as the prefill bulk). Streaming thus trades a small fixed exact buffer plus
extra \emph{decode-time} SVDs (one per compressed chunk) for its $\mathcal{O}(g)$ tail, winning
when $g$ is large. The chunk $\ell$ sets the granularity. It must exceed the rank for the
compression to be lossy, and larger $\ell$ relative to $d$ tightens the storage toward its $r/d$
asymptote; taking $w\!\to\!\infty$ recovers the accumulating cache (further details in Table~\ref{tab:mem}).
We use $\ell{=}128$ in our experiments, above every rank we use so each aged-out chunk is genuinely truncated, and a larger $\ell$ is equally valid.

\subsection{Interpretability: what the retained basis exposes}
\label{sec:interp}
\keybox{\textcolor{ourframe}{\textbf{Insight.}} By compressing in the score and output
space, weighting $K$ by the query covariance $C_Q{=}Q^\top Q$ and $V$ by the per-token
attention mass, \method\ retains a head's \emph{functional} subspace. Because this basis is derived
from the prompt's own attention geometry rather than from fixed weights or calibration data, these
same quantities also serve as free, per-head interpretability insights: they reveal
how each head geometrically uses its context, rather than what it attends to semantically.}
\emph{(i)~Effective rank.} \citet{roy} observe that the effective rank of a matrix such as $K$ or
$V$ measures its \emph{intrinsic dimensionality}. Because our compression happens in the score and
output geometry, the
per-matrix ranks $r_{\widetilde K},r_{\widetilde V}$ of Eq.~\eqref{eq:energyrank} report the intrinsic
dimensionality of the scores $P$ (through the query-weighted keys $\widetilde K$) and of the output
$O$ (through the mass-weighted values $\widetilde V$). The per-matrix ranks can be read as the
intrinsic compression capacity of each matrix, and the per-(layer, head) rank map shows which heads
carry low-rank, highly compressible context. We empirically observe that, on the models we test, the
query- and mass-weightings redistribute singular-value energy yet still leave a systematic gap
$r_{\widetilde K}\!\ll\!r_{\widetilde V}$ (Appendix~\ref{app:extexp},
Figures~\ref{fig:svd-anatomy} and~\ref{fig:svd-anatomy-v}).

\emph{(ii)~Query subspace.} The covariance $C_Q=Q^\top Q$ is the subspace the head's queries actually
probe; the decay of its spectrum quantifies how \emph{focused} (few-directional) vs.\ diffuse a
head's attention is.

\emph{(iii)~Token saliency.} The per-key mass $a_i=\sum_q P_{qi}$ is a token-importance profile (the
same heavy-hitter signal exploited by eviction~\citep{h2o}) emerged here freely, so even an
eviction-only pipeline can read off which tokens to keep and which to drop.

We emphasize that this is \emph{geometric}, not \emph{semantic}. The basis vectors are
singular directions, not named concepts, they mix position and content (post-RoPE),
and they describe each prompt and head rather than explain a prediction. Within those limits, the
effective-rank map, query-subspace concentration, and attention-mass profile are concrete,
reusable quantities the compression produces at no cost (interpretability analysis in Appendix~\ref{app:extexp}).

\section{Experiments}

\paragraph{Baselines.}
\label{sec:baselines}
We compare \method\ against three reference points at a token budget matched to \method's memory at each compression setting, so
every method shares the same model, prompts, decoding, and cache size. The \emph{dense cache} is the
uncompressed upper bound. We compare against \emph{eviction} methods: \textbf{H2O} (heavy-hitter attention mass), \textbf{KeyDiff}
(attention-free cosine similarity), and \textbf{KVzip} (query-agnostic reconstruction
scoring); and \emph{low-rank compression} methods: \textbf{SVD on raw KV} (standard top-$r$ truncated SVD applied directly to $K$ and $V$, without \method's attention
weighting), \textbf{KQ-SVD} (attention-aware but from an \emph{offline calibration} set),
and \textbf{OjaKV} (an \emph{online} Oja's-rule basis seeded from calibration).

\paragraph{Datasets and benchmarks.} We evaluate on \emph{two complementary benchmarks}.

\emph{(i) Agentic benchmark}: $184$ prompts spanning $13$ common
scenarios across three difficulty tiers, each annotated with the correct action to take (from
${\sim}77$ candidates) or \emph{no} action when none applies, examining action \emph{selection}
and \emph{necessity}. An LLM judge scores each output $1$ to $5$; we
report the mean judge score and $\mathrm{pass@}4$ (fraction $\ge 4$). Prompts are short but the
reasoning-and-response generations are long (median ${\sim}800$ tokens, up to ${\sim}10$K).

\emph{(ii) LongBench}~\citep{bai2024longbench}: the full public long-context suite ($1{,}750$ examples over $9$ tasks),
each scored by its native metric; we report per-task accuracy and the average ($0$ to $100$).
Its long-context inputs and outputs stress the cache at real context length and exercise the
streaming schedule's sliding window. The two examine different axes: the agentic eval isolates answer
\emph{quality}, LongBench tests \emph{generality} at long context, so agreement across both guards
against over-reliance on either. Memory sums the \emph{actual} stored tensor bytes (factors plus
exact tail) from one prefill forward, cross-checked against the allocator's~peak (full setup and model choice in Appendix~\ref{app:ablations}).

\begin{table}[t]
\centering
\small
\vspace{-10pt}
\caption{Measured KV-cache memory on Phi-4-14B
(40 layers, 40 query / 10 KV heads) and Qwen3-8B (36 layers, 32 query / 8 KV heads); $d{=}128$, bf16.}
\vspace{5pt}
\resizebox{0.92\linewidth}{!}{%
\begin{tabular}{lccccccc}
\toprule
& & \multicolumn{3}{c}{Phi-4-14B} & \multicolumn{3}{c}{Qwen3-8B} \\
\cmidrule(lr){3-5}\cmidrule(lr){6-8}
& Rank $r$ & KV @32K (GiB) & Measured ratio & KV @9.4K (GiB) & KV @32K (GiB) & Measured ratio & KV @9.4K (GiB) \\
\midrule
\rowcolor{fullcol}
Dense cache      & $128$ & $6.250$ & $1.000$ & $1.804$ & $4.500$ & $1.000$ & $1.299$ \\
\method\ ($c{=}0.50$) & $64$  & $3.139$ & $0.502$ & $0.916$ & $2.260$ & $0.502$ & $0.660$ \\
\method\ ($c{=}0.75$) & $32$  & $1.571$ & $0.251$ & $0.459$ & $1.131$ & $0.251$ & $0.330$ \\
\method\ ($c{=}0.875$)& $16$  & $0.787$ & $0.126$ & $0.231$ & $0.567$ & $0.126$ & $0.166$ \\
\bottomrule
\end{tabular}%
}
\label{tab:mem}
\end{table}

\subsection{\method\ vs.\ eviction methods}
\label{sec:vs-eviction}
\begin{figure}[!t]
\centering
\includegraphics[width=\linewidth]{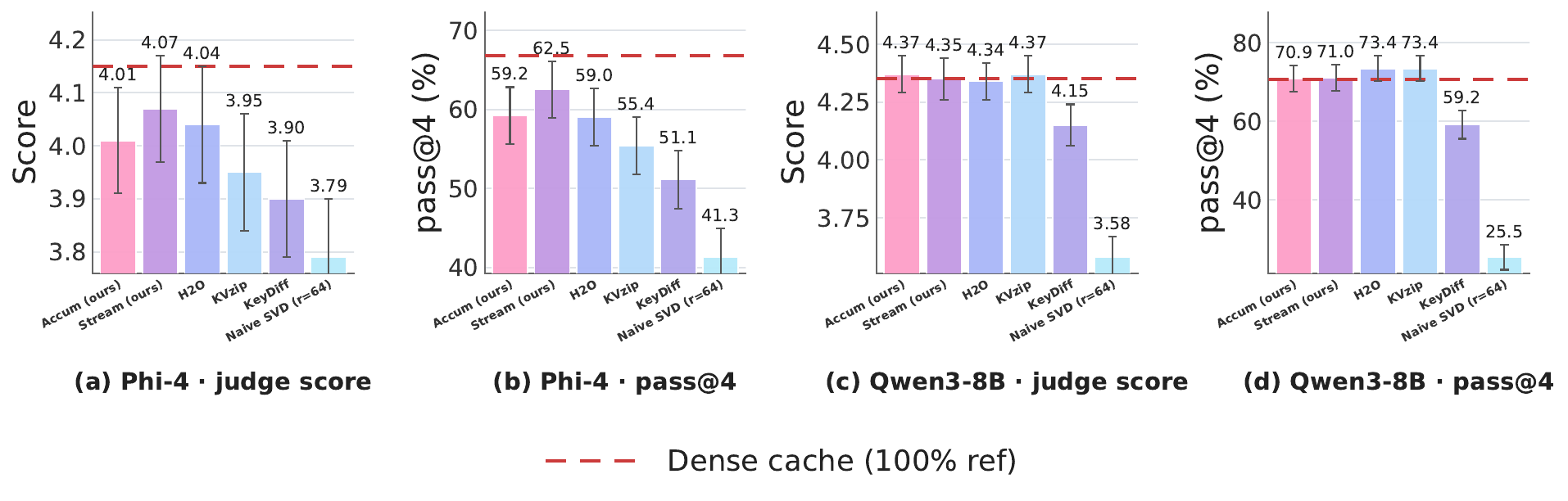}
\caption{At the aggressive half-rank ($50\%$ cache) point, \method\ stays within noise of the dense
cache and matches or beats the strongest baselines on both Phi-4-14B and Qwen3-8B.}
\label{fig:quality}
\end{figure}

\begin{figure}[!t]
\centering
\includegraphics[width=\linewidth]{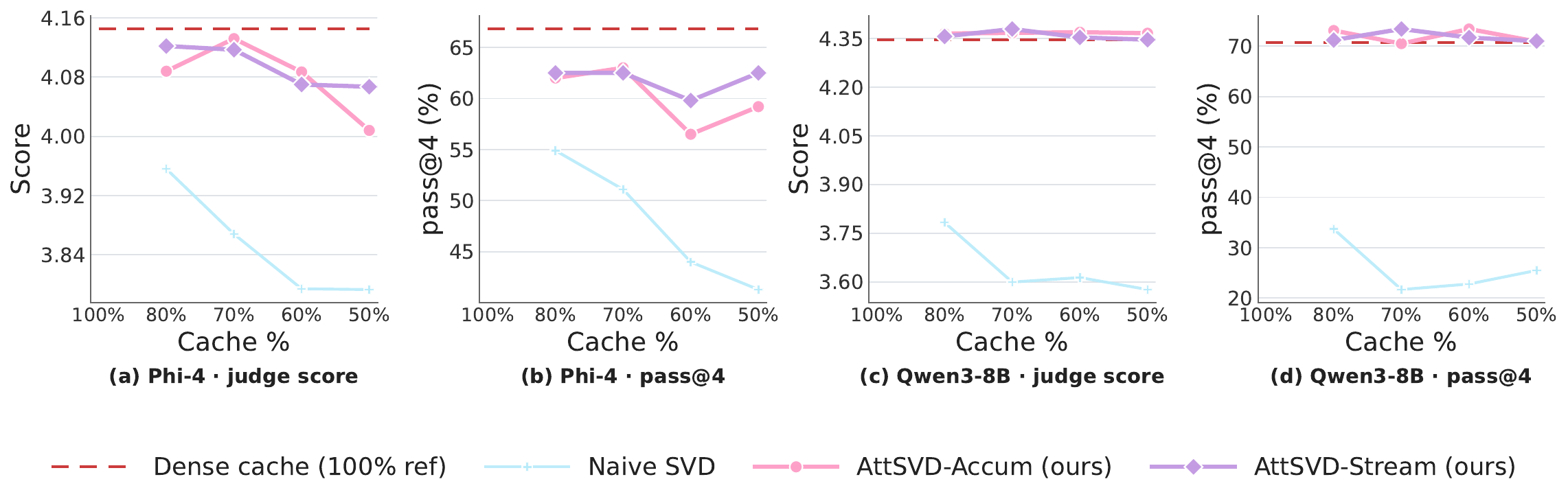}
\caption{\textbf{Agentic benchmark} ($n{=}184$): mean judge score (a,\,c) and $\mathrm{pass@}4$ (b,\,d)
as the cache is reduced from $100\%$ down to $50\%$, on Phi-4-14B (a,\,b) and Qwen3-8B (c,\,d).
\method\ (Accum/Stream) follows the dense cache far more closely
than SVD on raw KV across the rank range.}
\label{fig:phi4sweep}
\end{figure}

On the agentic benchmark at the
aggressive half-rank ($50\%$ cache) point (Figure~\ref{fig:quality}), attention-aware \method-Stream
reaches $4.07$ mean judge\,/\,$62.5\%$ $\mathrm{pass@}4$, edging out the strongest eviction baselines
(H2O $4.04$/$59.0\%$, KVzip $3.95$/$55.4\%$, KeyDiff $3.90$/$51.1\%$), while its \method-Accum cache
($4.01$/$59.2\%$) is on par with H2O at half the memory. On Qwen3-8B
(Figure~\ref{fig:quality}) the mean-judge gaps among Full, \method\ (Accum/Streaming) and the eviction
baselines H2O/KVzip all fall within the reported $95\%$ CI, they are statistically tied with full
attention, and only KeyDiff differs. The ordering is stable across the full rank range
(Figure~\ref{fig:phi4sweep}). The same holds for long context: on LongBench at the same $50\%$ cache
point (Table~\ref{tab:longbench}), \method-Accum leads every eviction baseline on Phi-4-14B ($36.0$ vs.\
KVzip $35.4$, H2O $34.9$, KeyDiff $33.5$) and ties the strongest eviction methods on Qwen3-8B
(KVzip $51.9$, H2O $51.2$), while staying within $0.7$ of the dense cache.

\begin{table}[!t]
\centering
\tiny
\vspace{-10pt}
\caption{At $50\%$ cache on LongBench, \method\ (Accum/Stream) matches the dense cache average on
both Phi-4-14B and Qwen3-8B and leads every reconstruction and eviction baseline.}
\vspace{5pt}
\setlength{\tabcolsep}{3pt}
\resizebox{0.92\linewidth}{!}{%
\begin{tabular}{l>{\columncolor{fullcol}}ccccccc@{\hspace{16pt}}>{\columncolor{fullcol}}ccccccc}
\toprule
 & \multicolumn{7}{c}{Phi-4-14B} & \multicolumn{7}{c}{Qwen3-8B} \\
\cmidrule(lr){2-8}\cmidrule(lr){9-15}
 & & \multicolumn{3}{c}{Reconstruction} & \multicolumn{3}{c}{Eviction} & & \multicolumn{3}{c}{Reconstruction} & \multicolumn{3}{c}{Eviction} \\
\cmidrule(lr){3-5}\cmidrule(lr){6-8}\cmidrule(lr){10-12}\cmidrule(lr){13-15}
LongBench task & Dense & \shortstack{\method-Accum\\{\tiny(ours)}} & \shortstack{\method-Stream\\{\tiny(ours)}} & SVD on raw KV & KeyDiff & KVzip & H2O & Dense & \shortstack{\method-Accum\\{\tiny(ours)}} & \shortstack{\method-Stream\\{\tiny(ours)}} & SVD on raw KV & KeyDiff & KVzip & H2O \\
\midrule
PassageRetrieval-en & $98.5$ & $98.5$ & $98.5$ & $99.0$ & $98.0$ & $98.5$ & $96.0$  & $100.0$ & $100.0$ & $100.0$ & $78.0$ & $100.0$ & $100.0$ & $99.5$ \\
Qasper              & $27.9$ & $26.9$ & $28.1$ & $21.8$ & $23.4$ & $28.1$ & $26.7$  & $41.0$ & $41.4$ & $41.4$ & $37.0$ & $30.7$ & $41.2$ & $37.5$ \\
MultiFieldQA-en     & $32.9$ & $33.0$ & $31.3$ & $23.4$ & $32.4$ & $29.6$ & $32.3$  & $53.8$ & $53.8$ & $53.8$ & $51.3$ & $44.9$ & $54.1$ & $51.7$ \\
HotpotQA            & $21.4$ & $20.7$ & $20.8$ & $14.8$ & $20.8$ & $20.5$ & $20.4$  & $59.4$ & $59.1$ & $59.1$ & $53.5$ & $56.8$ & $58.9$ & $58.3$ \\
2WikiMQA            & $36.5$ & $35.6$ & $35.8$ & $21.9$ & $27.2$ & $33.3$ & $33.8$  & $43.0$ & $43.3$ & $43.3$ & $38.1$ & $40.9$ & $44.0$ & $43.0$ \\
GovReport           & $14.3$ & $14.1$ & $14.2$ & $13.8$ & $14.0$ & $14.4$ & $14.6$  & $18.5$ & $18.6$ & $18.9$ & $16.8$ & $17.1$ & $18.6$ & $18.6$ \\
TriviaQA            & $28.6$ & $26.6$ & $26.8$ & $20.3$ & $26.8$ & $27.6$ & $27.3$  & $88.5$ & $89.0$ & $89.0$ & $87.1$ & $85.2$ & $88.5$ & $88.5$ \\
TREC                & $52.5$ & $53.3$ & $51.9$ & $52.3$ & $47.8$ & $58.5$ & $51.2$  & $59.5$ & $57.7$ & $57.7$ & $25.8$ & $35.1$ & $62.1$ & $61.1$ \\
PassageCount        & $17.5$ & $15.5$ & $14.0$ & $12.0$ & $11.0$ & $8.0$ & $12.0$  & $2.5$ & $3.5$ & $3.5$ & $2.5$ & $6.0$ & $0.0$ & $2.5$ \\
\midrule
average             & $36.7$ & $\mathbf{36.0}$ & $\underline{35.7}$ & $31.0$ & $33.5$ & $35.4$ & $34.9$  & $51.8$ & $51.8$ & $\mathbf{51.9}$ & $43.3$ & $46.3$ & $\underline{51.9}$ & $51.2$ \\
\bottomrule
\end{tabular}%
}

\label{tab:longbench}
\end{table}

\subsection{\method\ vs.\ other low-rank compression methods}
\label{sec:vs-svd}
Against other low-rank / SVD compressors the \emph{basis} is what matters, not just the rank. SVD on raw KV
truncation trails even the eviction family at
$r{=}64$ on the agentic benchmark ($3.79$/$41.3\%$, Figure~\ref{fig:quality}), reflecting that a
softmax-pretrained model has not adapted to low-rank $K/V$; attention-aware \method\ closes this gap
outright, adding $+0.13$ to $0.29$ mean judge and $+7$ to $18$ points $\mathrm{pass@}4$ over SVD on raw KV
at matched memory across the rank range (Figure~\ref{fig:phi4sweep}). The same holds on LongBench, where
\method\ ($36.0$) is the strongest reconstruction compressor and SVD on raw KV trails ($31.0$;
Table~\ref{tab:longbench}). We further test two \emph{calibrated} SVD baselines on a third backbone,
Llama-3.1-8B-Instruct,\footnote{We use Llama-3.1-8B-Instruct here because both baselines ship
official code for it. KQ-SVD evaluates on LLaMA/Mistral, and OjaKV uses Llama-3.1-8B as its reference model.}
(Table~\ref{tab:longbench-llama}): KQ-SVD and OjaKV. We run
each with its official implementation and released settings, matched only in cache budget, so the
numbers reflect the methods as released.
\method-Accum matches the dense cache almost exactly on the average ($48.5$ vs.\ $48.5$ at
$80\%$ cache; $47.8$ vs.\ $48.5$ at $60\%$) and leads KQ-SVD ($47.4$ and $45.2$) by $1.1$ to $2.6$
points, with the largest gaps on the QA tasks (HotpotQA, MultiFieldQA-en) where a fixed calibration
basis transfers least well across prompts. Both \method\ schedules lead the OjaKV
baseline ($42.9$ at $80\%$), which collapses to $30.5$ at $60\%$. These gaps show that
\method's per-prompt, attention-aware basis generalizes better than both KQ-SVD and OjaKV,
widening its advantage as the cache tightens.

\begin{table}[!t]
\centering
\tiny
\vspace{-10pt}
\caption{On Llama-3.1-8B-Instruct, \method\ matches the dense cache average at $80\%$ ($48.5$ vs.\
$48.5$) and stays within $0.7$ at $60\%$, beating KQ-SVD and OjaKV at both matched cache points.}
\vspace{5pt}
\setlength{\tabcolsep}{3pt}
\resizebox{0.68\linewidth}{!}{%
\begin{tabular}{l>{\columncolor{fullcol}}ccccc@{\hspace{14pt}}>{\columncolor{fullcol}}ccccc}
\toprule
 & \multicolumn{5}{c}{$80\%$ cache ($c{=}0.2$, $r{\approx}102$)} & \multicolumn{5}{c}{$60\%$ cache ($c{=}0.4$, $r{\approx}77$)} \\
\cmidrule(lr){2-6}\cmidrule(lr){7-11}
LongBench task & Dense & \shortstack{\method-Accum\\{\tiny(ours)}} & \shortstack{\method-Stream\\{\tiny(ours)}} & KQ-SVD & OjaKV & Dense & \shortstack{\method-Accum\\{\tiny(ours)}} & \shortstack{\method-Stream\\{\tiny(ours)}} & KQ-SVD & OjaKV \\
\midrule
MultiFieldQA-en     & $53.5$ & $53.3$ & $53.3$ & $50.5$ & $48.3$ & $53.5$ & $51.2$ & $51.2$ & $48.1$ & $45.6$ \\
HotpotQA            & $54.2$ & $54.3$ & $54.3$ & $50.9$ & $35.8$ & $54.2$ & $53.5$ & $53.5$ & $48.2$ & $26.9$ \\
2WikiMQA            & $44.6$ & $45.3$ & $45.3$ & $40.9$ & $40.1$ & $44.6$ & $45.4$ & $45.4$ & $40.4$ & $31.8$ \\
TREC                & $7.5$ & $7.0$ & $7.0$ & $12.0$ & $8.2$ & $7.5$ & $6.0$ & $6.0$ & $7.8$ & $3.0$ \\
PassageCount        & $6.2$ & $6.8$ & $6.8$ & $7.2$ & $3.0$ & $6.2$ & $7.2$ & $7.2$ & $7.6$ & $4.5$ \\
PassageRetrieval-en & $99.5$ & $99.5$ & $99.5$ & $99.5$ & $89.0$ & $99.5$ & $99.5$ & $99.5$ & $99.5$ & $36.2$ \\
Qasper              & $47.4$ & $47.6$ & $47.6$ & $46.5$ & $45.1$ & $47.4$ & $46.3$ & $46.3$ & $45.3$ & $37.7$ \\
GovReport           & $34.7$ & $33.8$ & $33.8$ & $32.2$ & $32.6$ & $34.7$ & $31.3$ & $31.3$ & $26.4$ & $26.3$ \\
TriviaQA            & $89.1$ & $89.2$ & $89.2$ & $86.6$ & $84.1$ & $89.1$ & $89.5$ & $89.5$ & $83.5$ & $62.2$ \\
\midrule
average             & $48.5$ & $\mathbf{48.5}$ & $\mathbf{48.5}$ & $47.4$ & $42.9$ & $48.5$ & $\mathbf{47.8}$ & $\mathbf{47.8}$ & $45.2$ & $30.5$ \\
\bottomrule
\end{tabular}%
}
\label{tab:longbench-llama}
\end{table}

\subsection{Dynamic energy-rule rank}
\label{sec:energy-results}
\begin{figure}[!t]
\centering
\includegraphics[width=0.55\linewidth]{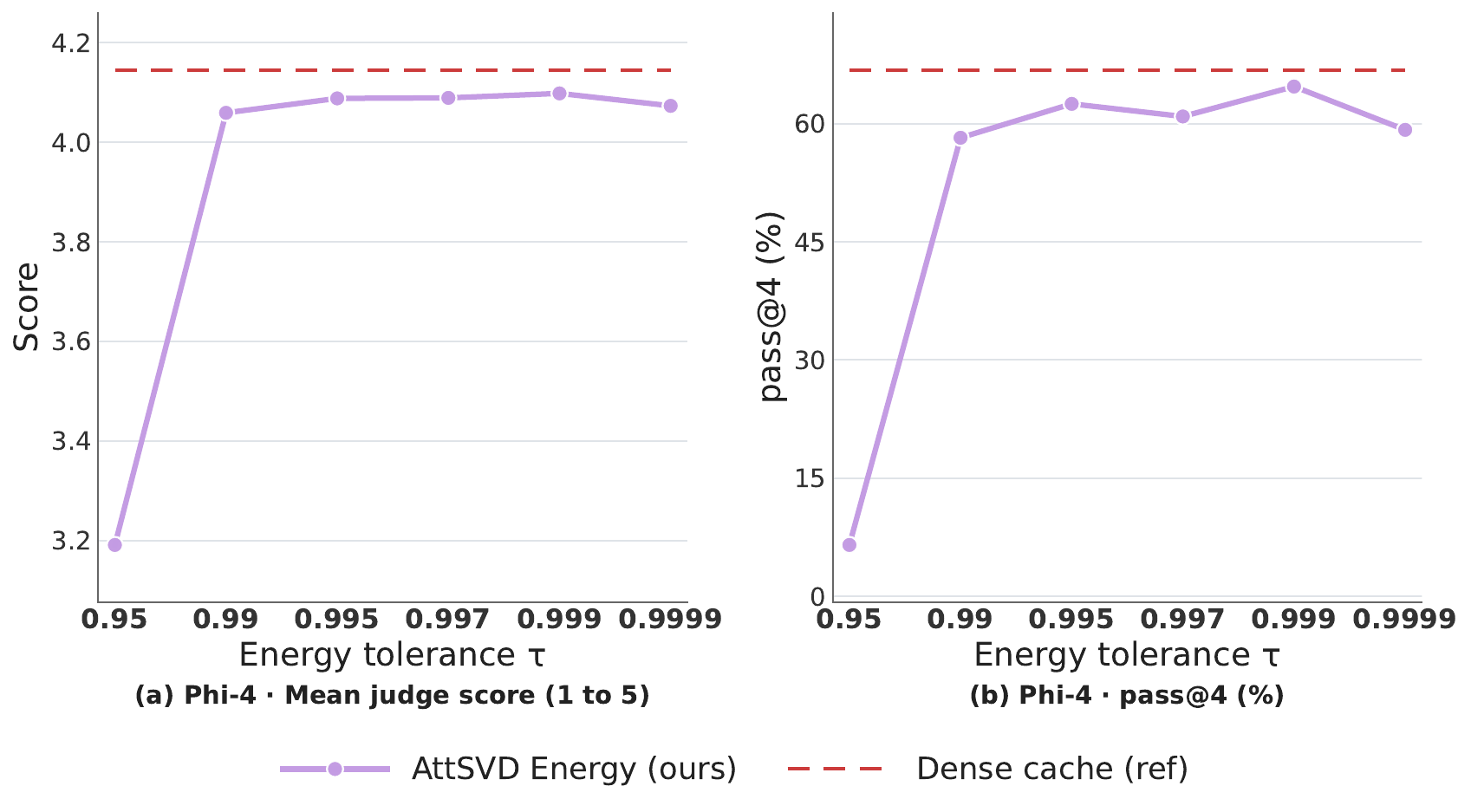}
\caption{Under the dynamic energy rule, quality saturates by $\tau{=}0.99$ and matches the dense
cache from there on (agentic benchmark, $n{=}184$).}
\label{fig:phi4energy}
\end{figure}

Section~\ref{sec:adaptive} sets $r_{\widetilde K},r_{\widetilde V}$ per prompt from each matrix's own spectrum at a single
tolerance $\tau$ (Eq.~\eqref{eq:energyrank}), with no fixed rank or ratio. On the agentic benchmark
(Figure~\ref{fig:phi4energy}) this dynamic mode matches the dense cache within $0.05$ mean judge by
$\tau{=}0.999$ ($4.098$/$64.7\%$ vs.\ $4.145$/$66.8\%$), picking $r_{\widetilde K},r_{\widetilde V}$ per prompt with neither a
rank nor a ratio preset. Running the same rule on the full LongBench suite (Phi-4-14B, attention-aware
basis; Table~\ref{tab:energyrank}), at $\tau{=}0.997$ it reads $r_{\widetilde K}\!\approx\!65$ and
$r_{\widetilde V}\!\approx\!127$ off the data, keys collapse to about half rank while values stay near-full, exactly
the $r_{\widetilde K}\!\ll\!r_{\widetilde V}$ asymmetry of Section~\ref{sec:adaptive}. At the operating point the data itself
selects ($\approx\!75\%$ cache), the average ($36.5$) lands within $0.2$ of the dense cache
($36.7$); the tolerance $\tau$ alone recovers near-exact quality while each prompt sets its own~rank.
For a detailed interpretability analysis on real models, see Appendix~\ref{app:extexp}.

\begin{table}[!t]
\centering
\tiny
\vspace{-10pt}
\caption{The dynamic per-matrix energy rule picks $r_{\widetilde K},r_{\widetilde V}$ from each
prompt's own spectrum at tolerance $\tau$ and reaches near-exact LongBench quality on Phi-4-14B and
Qwen3-8B.}
\vspace{5pt}
\setlength{\tabcolsep}{4pt}
\resizebox{0.6\linewidth}{!}{%
\begin{tabular}{l>{\columncolor{fullcol}}ccc>{\columncolor{fullcol}}ccc}
\toprule
 & \multicolumn{3}{c}{Phi-4-14B} & \multicolumn{3}{c}{Qwen3-8B} \\
\cmidrule(lr){2-4}\cmidrule(lr){5-7}
LongBench task & Dense & Energy $\tau{=}0.997$ & Energy $\tau{=}0.995$ & Dense & Energy $\tau{=}0.997$ & Energy $\tau{=}0.995$ \\
\midrule
PassageRetrieval-en & $98.5$ & $99.5$ & $98.5$ & $100.0$ & $100.0$ & $100.0$ \\
Qasper              & $27.9$ & $27.0$ & $28.7$ & $41.0$ & $40.5$ & $41.2$ \\
MultiFieldQA-en     & $32.9$ & $33.1$ & $31.2$ & $53.8$ & $53.8$ & $53.2$ \\
HotpotQA            & $21.4$ & $21.4$ & $22.7$ & $59.4$ & $59.3$ & $59.3$ \\
2WikiMQA            & $36.5$ & $35.8$ & $35.6$ & $43.0$ & $43.5$ & $43.3$ \\
GovReport           & $14.3$ & $14.4$ & $14.5$ & $18.5$ & $19.2$ & $19.2$ \\
TriviaQA            & $28.6$ & $28.3$ & $28.4$ & $88.5$ & $88.5$ & $88.5$ \\
TREC                & $52.5$ & $53.9$ & $52.9$ & $59.5$ & $60.1$ & $61.4$ \\
PassageCount        & $17.5$ & $15.5$ & $15.0$ & $2.5$ & $3.0$ & $3.5$ \\
\midrule
average             & $36.7$ & $\mathbf{36.5}$ & $\underline{36.4}$ & $51.8$ & $\underline{52.0}$ & $\mathbf{52.2}$ \\
\bottomrule
\end{tabular}%
}
\label{tab:energyrank}
\end{table}


\section{Conclusion}
\method\ shows that the KV cache can be compressed without discarding tokens: it retains every
token yet stores each in an instance-optimal rank-$r$ subspace obtained from an \emph{online,
per-prompt} SVD and reconstructed lazily at decode, so the persistent
cache shrinks to ${\approx}r/d$ of dense. Its contributions are threefold: an online, calibration-free
basis that is provably optimal (Eckart-Young) for the prompt at hand; adaptive per-matrix ranks and
an attention-aware objective that spend the budget where attention actually concentrates; and accumulating
and streaming decode modes for different generation scenarios. Because the factors are exact, they
are also directly \emph{interpretable}: the effective-rank map exposes the intrinsic
compressibility of the query-weighted keys $\widetilde K$ and mass-weighted values
$\widetilde V$, the query subspace shows each head's focus, and the attention-mass
profile flags salient tokens, all without calibration. Empirically,
across several models, \method\ matches the dense cache on the agentic benchmark and
LongBench at roughly half the KV-cache memory,
positioning instance-optimal reconstruction as a strong, training-free alternative to eviction.

\clearpage
\section*{Use of AI Tools}
All ideas presented in this paper are original to the authors, or build upon the
related work cited herein. AI-based tools were used only in an assistive capacity: to improve the clarity
of the authors' initial statements, to help draft sections of the paper in \LaTeX, to improve the quality
of figures and tables, to correct grammatical errors, to make sentences more salient, and to serve as a coding assistant that helps implement the authors' ideas. All technical content, claims, and conclusions
were conceived, verified, and approved by the authors.

\section*{Ethics Statement}
The authors confirm that this work adheres to established ethical standards for machine learning research.
\method\ is a training-free, inference-time KV-cache compression method. Because the method operates purely
on the key and value caches of a frozen pretrained model, it
introduces no new training signal; any bias or fairness limitations already present in the underlying
language models are inherited unchanged, and \method\ is designed to alter neither their behavior nor their
outputs beyond compressing the cache. The authors report no competing interests and no external funding that
could have influenced the findings.

\section*{Reproducibility Statement}
We have taken concrete steps to make each reported result reproducible. The method, including the
attention-guided per-prompt truncated SVD, the energy-based rank rule, and the accumulating and streaming
decode strategies, is specified in Section~\ref{sec:method}, with the full procedures given as explicit
pseudocode in Appendix~\ref{app:algorithms} (Algorithms~\ref{alg:main} and~\ref{alg:attn}) and the
attention-aware truncation derived with a complete proof in Appendix~\ref{app:kqproof}. The experimental
setup, including implementation details, baselines, models, and extended studies, is documented in
Appendix~\ref{app:ablations}. To further support exact reproduction, we
release the core code of \method, the model-independent dynamic cache class (a \texttt{DynamicCache}
subclass that implements Algorithms~\ref{alg:main} and~\ref{alg:attn}) together with its unit tests, as
anonymized supplementary material, and defer the release of the full repository to acceptance.

\bibliographystyle{iclr2027_conference}
\bibliography{references}

\clearpage
\appendix
\raggedbottom

\section{Extended related work}\label{app:related}

\paragraph{Eviction methods.}
\emph{Eviction} methods take an orthogonal route, keeping a token budget
and dropping the rest by scoring tokens on accumulated heavy-hitter attention mass
(H2O~\citep{h2o}), attention-free key cosine similarity (KeyDiff~\citep{keydiff}), or query-agnostic
reconstruction importance (KVzip~\citep{kvzip}); we compare against all three. A parallel line keeps
\emph{eviction} as the backbone but relaxes its binary keep/drop decision into a non-binary
retain\,/\,approximate\,/\,evict allocation (D2O~\citep{d2o}, ARKV~\citep{arkv}).

\paragraph{Compression and reconstruction methods.}
Most methods extract the basis ahead of time and reuse it for every prompt; Table~\ref{tab:related}
groups them by basis source. \emph{Weight decompositions} apply an offline SVD to the key/value
projection \emph{weight matrices} (ASVD~\citep{asvd}, Palu~\citep{palu}, LoRC~\citep{lorc}). \emph{Calibration} methods estimate
the basis from a corpus (Eigen~Attention~\citep{eigen}, ReCalKV~\citep{recalkv});
KQ-SVD~\citep{kqsvd} is closest in spirit, factoring the
\emph{query-key product} $KQ^\top$ to target the attention scores rather than $K$ alone, but still
from an \emph{offline} calibration set of queries fixed across prompts. \emph{Cross-layer or trained}
methods share or learn the basis (xKV~\citep{xkv}, MatryoshkaKV~\citep{matryoshkakv},
Stiefel-manifold projections~\citep{stiefel}, EliteKV~\citep{elitekv}). All of
these tie the basis to weights, a calibration distribution, or a
learning phase. ShadowKV~\citep{shadowkv} is per-prompt, applying an online SVD to the pre-RoPE keys
with no calibration, but it compresses keys only and offloads the full values to CPU. ThinK~\citep{think}
is likewise per-prompt, pruning key channels by query-driven scoring with no training (keys only).
OjaKV~\citep{ojakv} is also online and per-context, i.e., it
maintains a \emph{running} incremental-PCA basis
(Oja's rule), seeded from a small calibration set and re-orthonormalized during decoding, and uses
attention only to choose which tokens stay full-rank rather than to shape the basis, trading
exactness for within-generation adaptation. A separate cache-style line predicts or residual-codes
values from keys via a fixed map $V\!\approx\!KW$ fit offline (DeltaKV~\citep{deltakv}), keeping keys
and approximating values; these share our ``keep all positions'' philosophy but with a
\emph{fixed, generic} map. In the same reconstruction-and-compression vein, VECTOR~\citep{vector}
reconstructs evicted values from retained keys by least squares, and Attention
Matching~\citep{attnmatch} fits the compressed cache to match the original \emph{attention output}
rather than the raw $K,V$; orthogonally, adaptive KV \emph{quantization} lowers the bit-width of the
stored cache rather than its rank (AQUA-KV~\citep{aquakv}). These are complementary to \method, which
instead sets the per-prompt low-rank basis itself from this prompt's own attention.

\begin{table}[H]
\centering\scriptsize
\vspace{-10pt}
\caption{SVD / low-rank KV-cache methods by basis source. $^\dagger$ShadowKV low-rank-compresses keys and offloads full values to CPU.}
\vspace{5pt}
\resizebox{\linewidth}{!}{%
\begin{tabular}{lllcc}
\toprule
Method & AttentionSVDs & Basis source & Calib./train? & K\&V \\
\midrule
ASVD~\citep{asvd}            & K/V proj.\ weights        & offline, activation-aware & calib. & weights \\
Palu~\citep{palu}           & K/V proj.\ weights        & offline, one-time         & calib. & K\,\&\,V \\
LoRC~\citep{lorc}           & KV weight matrices        & offline, layer-progressive& calib. & K\,\&\,V \\
Eigen Attn~\citep{eigen}    & K/Q/V activations         & offline calibration       & calib. & K/Q/V \\
KQ-SVD~\citep{kqsvd}        & \emph{score} $KQ^\top$/$VW_O$ & offline calibration (queries)& calib. & K\,\&\,V \\
ReCalKV~\citep{recalkv}     & K (grouped) / V           & offline calibration       & calib. & K\,\&\,V \\
ShadowKV~\citep{shadowkv}   & pre-RoPE \emph{keys}      & per-prompt SVD (keys only)& none   & K only$^\dagger$ \\
xKV~\citep{xkv}             & KV across layers          & post-hoc, shared          & none   & K\,\&\,V \\
MatryoshkaKV~\citep{matryoshkakv} & KV feature dim        & \emph{trained} projections& train  & K\,\&\,V \\
ThinK~\citep{think}         & key \emph{channels}       & query-aware (axis-aligned)& none   & K only \\
OjaKV~\citep{ojakv}         & K/V activations           & \emph{online} Oja PCA (seeded)& seed & K\,\&\,V \\
\midrule
\textbf{\method\ (ours)}   & $KC_Q^{1/2}$, $\mathrm{diag}(\sqrt a)V$, post-RoPE, per prompt & \emph{online} exact per-prompt SVD & \textbf{none} & K\,\&\,V \\
\bottomrule
\end{tabular}%
}
\label{tab:related}
\end{table}


\section{Proof of Eq.~\eqref{eq:kqmetric} (attention-aware key truncation)}
\label{app:kqproof}

We show that the query-metric key objective of Eq.~\eqref{eq:kqmetric} is (i) identical to a plain
Frobenius objective in the metric $C_Q$, and (ii) solved exactly by an ordinary truncated SVD after
a change of variables, yielding the factors $A_{\widetilde K},B_{\widetilde K}$ stored in Algorithm~\ref{alg:attn}. Throughout,
$Q\in\R^{m\times d}$ are the (post-RoPE) queries, $K\in\R^{n\times d}$ the keys,
$C_Q=Q^\top Q\in\R^{d\times d}$, and $E=K-X$ for a candidate rank-$r$ approximation $X$. As
$C_Q\succeq0$ is symmetric positive semidefinite it admits a unique symmetric PSD square root
$C_Q^{1/2}$; with the ridge $C_Q=Q^\top Q+\epsilon\,\bar c\,I_d\succ0$~\citep{hoerl1970ridge} used in
Algorithm~\ref{alg:attn}, $C_Q^{1/2}$ is invertible with inverse $C_Q^{-1/2}$.

\paragraph{(i) Norm equivalence.} Using $\lVert M\rVert_F^2=\mathrm{tr}(MM^\top)$ and the cyclic
invariance of the trace,
\begin{align*}
\big\lVert Q E^\top\big\rVert_F^2
 &= \mathrm{tr}\!\big(Q E^\top (Q E^\top)^\top\big)
  = \mathrm{tr}\!\big(Q E^\top E\, Q^\top\big)
  = \mathrm{tr}\!\big(E^\top E\, Q^\top Q\big)
  = \mathrm{tr}\!\big(E^\top E\, C_Q\big),\\[2pt]
\big\lVert E\, C_Q^{1/2}\big\rVert_F^2
 &= \mathrm{tr}\!\big(E\, C_Q^{1/2} C_Q^{1/2} E^\top\big)
  = \mathrm{tr}\!\big(E\, C_Q\, E^\top\big)
  = \mathrm{tr}\!\big(E^\top E\, C_Q\big),
\end{align*}
where both chains use cyclicity of the trace and the second uses $C_Q^{1/2}C_Q^{1/2}=C_Q$. Hence
$\lVert Q(K-X)^\top\rVert_F=\lVert (K-X)\,C_Q^{1/2}\rVert_F$ for every $X$, i.e.\ the two objectives
in Eq.~\eqref{eq:kqmetric} coincide; both equal $\mathrm{tr}\!\big((K-X)\,C_Q\,(K-X)^\top\big)^{1/2}$.
Up to the constant $1/\sqrt d$ this is precisely the perturbation of the logits $Q(K-X)^\top/\sqrt d$,
so minimizing the reconstruction error of $K$ in the $C_Q$ metric minimizes the logit error. Weighting
the factorization by the query-side second moment $C_Q=Q^\top Q$, rather than by a fixed calibration
basis, is the metric analogue of learning a low-rank attention subspace~\citep{eigen}.

\paragraph{(ii) Generalized Eckart-Young.} Introduce $\widetilde K=K\,C_Q^{1/2}$ and
$\widetilde X=X\,C_Q^{1/2}$. Since $C_Q^{1/2}$ is invertible, $\mathrm{rank}(\widetilde X)=\mathrm{rank}(X)$,
so $X\mapsto\widetilde X$ is a bijection on the set of rank-$\le r$ matrices, and
$(K-X)C_Q^{1/2}=\widetilde K-\widetilde X$. The problem of Eq.~\eqref{eq:kqmetric} thus reduces to the
\emph{ordinary} rank-constrained Frobenius problem
\[
\min_{\mathrm{rank}(\widetilde X)\le r}\big\lVert \widetilde K-\widetilde X\big\rVert_F .
\]
By the Eckart-Young theorem~\citep{eckartyoung1936,mirsky1960} (Eq.~\eqref{eq:ey}) its minimizer is the rank-$r$ truncated SVD of
$\widetilde K$: writing $\widetilde K=U\Sigma\mathcal V^\top$ with singular values nonincreasing,
$\widetilde X^\star=U_{:,1:r}\Sigma_{1:r}\mathcal V_{:,1:r}^\top$. Mapping back through $C_Q^{-1/2}$,
\[
X^\star=\widetilde X^\star C_Q^{-1/2}
 =\underbrace{U_{:,1:r}\Sigma_{1:r}}_{A_{\widetilde K}}\;\underbrace{\mathcal V_{:,1:r}^\top C_Q^{-1/2}}_{B_{\widetilde K}},
\]
which is exactly the decomposed store $A_{\widetilde K} B_{\widetilde K}=K_r^\star$ of Eq.~\eqref{eq:kqmetric} and
Algorithm~\ref{alg:attn}. Because the change of variables preserves the objective value and is a
bijection, $X^\star$ is a global minimizer of the original query-metric problem; it is unique whenever
$\sigma_r(\widetilde K)>\sigma_{r+1}(\widetilde K)$. \hfill$\square$

\paragraph{Remark (values, Eq.~\eqref{eq:vmass}).} The value objective is the same statement with a
left (row) metric $\mathrm{diag}(a)$ replacing the right (column) metric $C_Q$: for $E=V-X$,
$\sum_i a_i\lVert E_i\rVert_2^2=\lVert\mathrm{diag}(\sqrt a)\,E\rVert_F^2$, so by the identical
change-of-variables argument the minimizer is the truncated SVD of $\mathrm{diag}(\sqrt a)\,V$ with the
left factor rescaled by $\mathrm{diag}(1/\sqrt a)$, giving $A_{\widetilde V},B_{\widetilde V}$ as in Algorithm~\ref{alg:attn}.

\paragraph{Why the diagonal (per-key mass) metric.} We deliberately replace the exact row metric
$P^\top P$ by $\mathrm{diag}(a)$, $a_i=\sum_q P_{qi}$, for four reasons; all stem from $P^\top P$
being an $S\times S$ object that grows with the context length, whereas the key metric $C_Q$ stays a
fixed $d\times d$ gram.
\emph{(i)~Cost.} $P^\top P$ is quadratic in $S$ to form and its symmetric square root is an
$\mathcal{O}(S^3)$ eigendecomposition, per layer and forward pass; at $S{=}32$K that is a
${\sim}4$\,GB, ${\sim}10^{13}$-flop object per head, against $\mathcal{O}(S)$ to form, store, and
invert $a$. A metric that costs more than the low-rank cache saves would defeat the method.
\emph{(ii)~No materialized $P$.} $P^\top P$ needs the dense $S\times S$ attention matrix, which
memory-efficient kernels (FlashAttention) never form; the mass $a=\sum_q P_{q,:}$ is one
attention-shaped reduction that accumulates incrementally as tokens arrive, so it stays compatible
with both the memory goal and standard kernels.
\emph{(iii)~Conditioning.} $P^\top P$ has rank at most the number of query rows and is singular
whenever queries are fewer than keys, so $(P^\top P)^{-1/2}$ is ill-defined on its null space;
$\mathrm{diag}(a)$ with the floor $a\!\leftarrow\!a+10^{-3}\bar a$ is strictly positive and trivially
invertible.
\emph{(iv)~First-order sufficiency.} Attention mass concentrates on a few heavy tokens (the
first-token sink alone carries a large share; Fig.~\ref{fig:svd-anatomy-v}), so the diagonal already
encodes which rows the output reads, while the off-diagonal terms of $P^\top P$ are second-order.
Empirically this diagonal weighting alone lowers the value output error
$\lVert P(V-V_r)\rVert_F/\lVert PV\rVert_F$ from $0.154$ to $0.012$ at $50\%$ cache, most of the
attainable gain.

\section{Algorithms}
\label{app:algorithms}

\FloatBarrier
\vspace{-8pt}
\begin{algorithm}[H]
\caption{\textsc{\method}; online per-prompt low-rank KV cache (one layer, one key/value head). A single
\textsc{Decode} serves both schedules: the \texttt{streaming} flag switches on decode-time compression
(branch-specific steps for both schedules are \textcolor{algstream}{highlighted in blue}).}
\label{alg:main}
\scriptsize
\begin{algorithmic}[1]
\Require keys $K$, values $V\in\R^{S\times d}$;\, post-RoPE queries $Q$ (only for the attention-aware basis);\, recent window $w$;\, \texttt{streaming} flag with chunk $\ell$
\Ensure factors $(A_{\widetilde K},B_{\widetilde K}),(A_{\widetilde V},B_{\widetilde V})$;\, exact tail $(T_K,T_V)$ (a sliding window when \texttt{streaming});\, decode blocks $(\mathcal D_K,\mathcal D_V)$
\Statex
\Statex \textbf{Rank rule} (the one knob, fixed once): \emph{fixed} $r{=}\mathrm{round}\big((1{-}c)d\big)$ for both $K,V$, \emph{or} \emph{per-matrix energy} $\tau$ (set $r{=}\textsc{None}$, so $r_K,r_V$ are read from each matrix's own spectrum, giving $r_K\!\ll\!r_V$). The prompt bulk is factored attention-aware (\textsc{AttentionSVD}, Alg.~\ref{alg:attn}); streaming decode chunks, which carry no queries, use the unweighted \textsc{TruncSVD}.
\Statex
\Procedure{Prefill}{$K,V,Q$}\Comment{once, after the prompt is encoded}
  \State $r \gets \mathrm{round}\!\big((1-c)\,d\big)$ \textbf{if} ratio $c$ is given \textbf{else} \textsc{None} \Comment{fixed unified rank, else per-matrix energy rule}
  \State $K^{(b)},T_K \gets K_{1:S-w},\,K_{S-w+1:S}$;\;\; $V^{(b)},T_V \gets V_{1:S-w},\,V_{S-w+1:S}$ \Comment{split bulk / exact tail}
  \State $(A_{\widetilde K},B_{\widetilde K}),(A_{\widetilde V},B_{\widetilde V}) \gets \textsc{AttentionSVD}(K^{(b)},V^{(b)},Q,r)$ \Comment{attention-aware factors (Alg.~\ref{alg:attn}); discard the bulk}
  \State $\mathcal D_K,\mathcal D_V \gets (\,),(\,)$ \Comment{decode-block lists; populated only when \texttt{streaming}}
  \State \Return $K,\,V$ \Comment{prefill served exactly from the original $K,V$}
\EndProcedure
\Statex
\Procedure{Decode}{$k,v$}\Comment{every generated token}
  \State $T_K \gets [\,T_K;\,k\,],\quad T_V \gets [\,T_V;\,v\,]$ \Comment{append the new token exactly}
  \If{\textcolor{algstream}{\texttt{streaming}}}\Comment{\textcolor{algstream}{\textbf{STREAMING}: factor aged-out tokens during decode}}
    \While{$\mathrm{rows}(T_K) - w \ge \ell$}\Comment{\textcolor{algstream}{exact window overflowed by a full chunk}}
      \State $C_K,C_V \gets$ oldest $\ell$ rows of $T_K,T_V$ \Comment{\textcolor{algstream}{still exact values}}
      \State $(a_K,b_K)\gets\textsc{TruncSVD}(C_K,r)$, $(a_V,b_V)\gets\textsc{TruncSVD}(C_V,r)$; append to $\mathcal D_K,\mathcal D_V$ \Comment{\textcolor{algstream}{aged-out tokens carry no queries $\Rightarrow$ unweighted SVD; freeze}}
      \State drop those oldest $\ell$ rows from $T_K,T_V$ \Comment{\textcolor{algstream}{window stays in $[w,\,w{+}\ell]$}}
    \EndWhile
  \Else\Comment{\textcolor{algstream}{\textbf{STATIC}: no decode-time compression}}
    \State \textcolor{algstream}{keep $T_K,T_V$ exact} \Comment{\textcolor{algstream}{the tail grows by this token; prefill factors fixed, $\mathcal D_K,\mathcal D_V$ stay empty}}
  \EndIf
  \State $K_{\mathrm{full}} \gets [\,A_{\widetilde K}B_{\widetilde K};\,\textsc{Recon}(\mathcal D_K);\,T_K\,]$;\;\; $V_{\mathrm{full}} \gets [\,A_{\widetilde V}B_{\widetilde V};\,\textsc{Recon}(\mathcal D_V);\,T_V\,]$ \Comment{bulk $\oplus$ decode blocks $\oplus$ exact tail/window}
  \State \Return $K_{\mathrm{full}},\,V_{\mathrm{full}}$ \Comment{accumulating: $\mathcal D$ empty $\Rightarrow[A_{\widetilde K}B_{\widetilde K};T_K]$;\, freed after this layer}
\EndProcedure
\Statex
\State \textsc{Recon}$(\mathcal D)$: concatenate $AB$ over all blocks $(A,B)\in\mathcal D$ (oldest first); empty in accumulating mode. \textsc{AttentionSVD} is the attention-aware factorization of Alg.~\ref{alg:attn}.
\end{algorithmic}
\end{algorithm}

\begin{algorithm}[H]
\caption{\textsc{AttentionSVD}: the attention-aware low-rank factorization for the prompt bulk
(one layer/head). Keys are truncated in the query geometry $C_Q$ and values in the attention-mass
metric, so the kept directions are exactly the ones attention reads. \textsc{TruncSVD} is the
rank-$r$ thin SVD (rank rule).}
\label{alg:attn}
\scriptsize
\begin{algorithmic}[1]
\Function{AttentionSVD}{$K,V,Q,r$}\Comment{attention-aware basis (our method); $r$ fixed, or \textsc{None}$\Rightarrow$ energy rule}
  \If{$Q$ unavailable}\Comment{rare: capture hook did not fire $\Rightarrow$ no weighting available}
    \State \Return $\textsc{TruncSVD}(K,r),\ \textsc{TruncSVD}(V,r)$ \Comment{degenerate unweighted fallback only (Eq.~\ref{eq:factors})}
  \EndIf
  \phase{Keys}\, \emph{truncate in the query geometry $C_Q$, preserving the logits $QK^\top$}
  \State $C_Q \gets Q^\top Q + \epsilon\,\bar c\,I_d,\ \ \bar c{=}\tfrac1d\mathrm{tr}(Q^\top Q)$;\;\; $W,\,W^{-1} \gets C_Q^{1/2},\ C_Q^{-1/2}$ \Comment{$d{\times}d$ gram; ridge $\epsilon$ (def.\ $10^{-3}$)}
  \State $(A_{\widetilde K},B_{\widetilde K}) \gets \textsc{TruncSVD}(KW,\,r);\quad B_{\widetilde K} \gets B_{\widetilde K}\,W^{-1}$ \Comment{$A_{\widetilde K}B_{\widetilde K}=K^\star_r$ (Eq.~\ref{eq:kqmetric})}
  \phase{Values}\, \emph{weight rows by attention mass $a$, preserving the output $PV$}
  \State $a \gets \textstyle\sum_q \mathrm{softmax}(QK^\top/\sqrt d)_{q,:} + 10^{-3}\,\bar a$ \Comment{per-key mass (GQA-expanded), floored}
  \State $(A_{\widetilde V},B_{\widetilde V}) \gets \textsc{TruncSVD}(\mathrm{diag}(\sqrt a)\,V,\,r);\quad A_{\widetilde V} \gets \mathrm{diag}(1/\sqrt a)\,A_{\widetilde V}$ \Comment{$A_{\widetilde V}B_{\widetilde V}=V^\star_r$ (Eq.~\ref{eq:vmass})}
  \State \Return $(A_{\widetilde K},B_{\widetilde K}),\,(A_{\widetilde V},B_{\widetilde V})$
\EndFunction
\Statex
\Function{TruncSVD}{$X,r$}\Comment{rank-$r$ thin SVD with the rank rule (linear-algebra primitive)}
  \State $U,\Sigma,\mathcal V \gets \svd\big(\mathrm{fp32}(X)\big)$ \Comment{singular values $\Sigma$ in descending order}
  \If{$r=\textsc{None}$}\Comment{dynamic per-matrix rank from $X$'s own spectrum}
    \State $r \gets \min\big\{\rho:\ \textstyle\sum_{i\le\rho}\Sigma_{ii}^2 \ \ge\ \tau\sum_i\Sigma_{ii}^2\big\}$ \Comment{energy rule (Eq.~\ref{eq:energyrank}); default $\tau{=}0.99$}
  \EndIf
  \State \Return $\big(\,U_{:,1:r}\Sigma_{1:r}\,,\ \ \mathcal V_{:,1:r}^\top\,\big)$ \Comment{$\in\R^{n\times r}\!\times\R^{r\times d}$; Eckart-Young (Eq.~\ref{eq:ey})}
\EndFunction
\end{algorithmic}
\end{algorithm}
\FloatBarrier


\section{Experimental setup and extended experiments}
\label{app:ablations}
\label{app:extexp}

On a small model we verify that full-rank ($r{=}d$) decomposition reproduces the
dense-attention output \emph{bit-for-bit}; the reconstruction error then
decreases monotonically as $r$ increases, consistent with Eq.~\eqref{eq:ey}. Table~\ref{tab:mem}
reports the KV-cache footprint on Phi-4-14B: the measured ratio matches the theoretical $r/d$ to the
decimal at $32$K context; at shorter context the fixed $rd$ and exact-tail terms make the ratio
slightly larger, exactly as Eq.~\eqref{eq:ratio} predicts.

\subsection{Streaming vs.\ accumulating decode}
\label{sec:vs-streaming}
The accumulating and streaming schedules differ only in the decode tail. Accumulating keeps every
earlier decode token exact (an $\mathcal{O}(g)$ tail that grows with the generation), while streaming
factors aged-out tokens into frozen rank-$r$ blocks so the exact buffer stays bounded. The
agentic benchmark generations are mostly too short to overflow the $w$-token window, so streaming rarely
evicts; to stress it where it genuinely compresses decode history we repeat the comparison on the
full LongBench suite ($n{=}1{,}750$), whose long-context outputs routinely overflow the window and
trigger repeated chunk compression (Table~\ref{tab:streaming-lb}). Across all ranks, streaming stays
within $0.3$ points of the accumulating (decode-exact) cache on average ($\Delta\in[-0.29,+0.24]$), a
gap well inside run-to-run noise, and it follows the dense cache ($36.7$) down to the $50\%$ budget.
This holds \emph{even though} streaming repeatedly factors aged-out tokens into frozen low-rank blocks
rather than keeping them exact. Across both a short-output agentic eval (Figure~\ref{fig:phi4sweep}) and
a long-output academic suite, then, bounding the $\mathcal{O}(g)$ decode tail to a fixed window costs
no measurable quality. Per task, the two schedules are in fact identical on $8$ of $9$ LongBench tasks
and differ only on GovReport, the sole task with a long ($512$-token) generation budget, where
streaming compression fires once the generation overflows the $w{=}64$ window; the remaining tasks
generate $\le\!128$ tokens (mostly $\le\!64$) and end early under greedy decoding, so the streaming
path never fires and reduces exactly to accumulating. At the aggressive
$r{=}64$ the streaming cache holds $62.5\%$ $\mathrm{pass@}4$ with the $w{=}64$ window, vs.\ $56.0\%$
at $w{=}16$ in an earlier run; enlarging the exact window removes the only point where compressing
decode history had cost quality. The result is decisive in one direction: streaming
\emph{loses nothing} relative to the accumulating cache, while replacing the accumulating cache's
$\mathcal{O}(g)$ exact tail (which grows with every one of the $g$ generated tokens) with a
\emph{bounded} one; a large memory-efficiency gain on long generations at no quality cost.

\begin{table}[!t]
\centering
\tiny
\vspace{-10pt}
\caption{Streaming vs.\ accumulating decode on LongBench (Phi-4-14B, full suite, $n{=}1{,}750$; average,
$\Delta$ the streaming-minus-accumulating gap), showing that streaming stays within run-to-run noise
of the accumulating cache at every budget and lets the growing $\mathcal{O}(g)$ exact decode tail be
bounded to a fixed window at no quality cost.}
\vspace{5pt}
\resizebox{\linewidth}{!}{%
\begin{tabular}{lllcccccccc}
\toprule
 & & & \multicolumn{4}{c}{Phi-4-14B} & \multicolumn{4}{c}{Qwen3-8B} \\
\cmidrule(lr){4-7}\cmidrule(lr){8-11}
Cache & $c$ & $r$ & SVD on raw KV & \shortstack{\method-Accum (decode exact)\\{\tiny(ours)}} & \shortstack{\method-Stream ($w{=}64$)\\{\tiny(ours)}} & $\Delta$ & SVD on raw KV & \shortstack{\method-Accum\\{\tiny(ours)}} & \shortstack{\method-Stream\\{\tiny(ours)}} & $\Delta$ \\
\midrule
\rowcolor{fullcol}
$100\%$ & $-$   & Dense cache & $-$ & \multicolumn{2}{c}{$36.7$} & $-$ & $-$ & \multicolumn{2}{c}{$51.8$} & $-$ \\
$80\%$  & $0.2$ & $102$ & $33.83$ & $36.21$ & $36.45$ & $+0.24$ & $48.84$ & $51.75$ & $51.75$ & $+0.00$ \\
$70\%$  & $0.3$ & $90$  & $33.13$ & $37.21$ & $36.93$ & $-0.28$ & $48.33$ & $52.10$ & $52.09$ & $-0.01$ \\
$60\%$  & $0.4$ & $77$  & $33.16$ & $36.64$ & $36.65$ & $+0.01$ & $46.88$ & $52.55$ & $52.54$ & $-0.01$ \\
$50\%$  & $0.5$ & $64$  & $31.0$ & $36.02$ & $35.73$ & $-0.29$ & $43.34$ & $51.81$ & $51.85$ & $+0.04$ \\
\bottomrule
\end{tabular}%
}
\label{tab:streaming-lb}
\end{table}

\subsection{Streaming-window ablation on a long-generation task}\label{app:streamwin}
\begin{table}[H]
\centering
\tiny
\vspace{-10pt}
\caption{Streaming-window ablation on GovReport (Llama-3.1-8B-Instruct, $c{=}0.2$)
for different exact decode-tail windows $w$, where ROUGE-L stays flat as $w$ shrinks from $\infty$
down to $16$ so even a small exact window preserves quality.}
\vspace{5pt}
\begin{tabular}{lcc}
\toprule
Decode schedule & Exact decode window $w$ & GovReport (ROUGE-L) \\
\midrule
\rowcolor{fullcol}
FullKV (parity anchor)      & n/a        & $34.66$ \\
\method-Accum (decode exact) & $\infty$ & $33.79$ \\
\method-Stream              & $256$      & $33.80$ \\
\method-Stream              & $128$      & $33.81$ \\
\method-Stream              & $64$       & $33.77$ \\
\method-Stream              & $32$       & $33.88$ \\
\method-Stream              & $16$       & $33.74$ \\
\bottomrule
\end{tabular}
\label{tab:streamwin}
\end{table}

The streaming schedule bounds the exact \emph{decode} tail to a sliding window of $w$ tokens,
factoring aged-out tokens into frozen low-rank blocks (\S\ref{sec:streaming}). On standard
LongBench QA this path rarely fires, because most tasks generate $\le\!128$ tokens and never
overflow the window, so streaming reduces to the accumulating (decode-exact) cache
(\S\ref{sec:baselines}; Table~\ref{tab:longbench-llama}). To isolate the effect of $w$ where the
schedule \emph{does} fire, we run a single-job ablation on GovReport, the LongBench task with the
longest generation budget ($512$ tokens), at a fixed cache ratio ($c{=}0.2$, i.e.\ $80\%$),
varying the window $w$ and including the accumulating cache as the $w{=}\infty$ limit. All variants share
the same attention-aware prefill factorization and are scored together against the FullKV anchor.
The result is a flat curve where ROUGE-L varies by under $0.15$ across $w$ from $256$ down to $16$ and
stays level with the $w{=}\infty$ accumulating cache, so bounding the exact decode window to as few as
$16$ tokens costs no measurable quality.


\subsection{Interpretability analysis}
\label{sec:ablation}
\afterpage{%
\begin{table}[t]
\centering
\tiny
\vspace{-10pt}
\caption{SVD on raw KV vs.\ attention-aware truncation at matched rank on a synthetic head whose
query-relevant directions are misaligned with $K$'s dominant singular directions. At equal rank
the attention-aware basis drives the attention-relevant error far below SVD on raw KV for both $K$ and
$V$, and reaches it exactly once the retained rank spans the weighted subspace.}
\vspace{5pt}
\begin{tabular}{lccc}
\toprule
Attention-relevant error (matched rank) & Rank $r$ & SVD on raw KV & Attention-aware \\
\midrule
$\lVert Q(K-K_r)^\top\rVert_F\,/\,\lVert QK^\top\rVert_F$ & $4$  & $0.976$ & $\mathbf{0.267}$ \\
$\lVert P(V-V_r)\rVert_F\,/\,\lVert PV\rVert_F$           & $8$  & $0.923$ & $\mathbf{0.643}$ \\
$\lVert P(V-V_r)\rVert_F\,/\,\lVert PV\rVert_F$           & $16$ & $0.608$ & $\mathbf{0.000}$ \\
\bottomrule
\end{tabular}
\label{tab:ablation}
\end{table}%
}

\subsubsection{Adversarial case on a controlled synthetic head}

We begin with a controlled synthetic head that isolates the core effect. Table~\ref{tab:ablation} is
built so that $K$'s query-relevant directions are deliberately misaligned with its dominant singular
directions, the adversarial case for energy-ordered SVD. At matched rank we compare SVD on raw KV,
which minimizes $\lVert K-K_r\rVert_F$, against the \emph{attention-aware} truncation of
Eqs.~\eqref{eq:kqmetric} and~\eqref{eq:vmass}. On the keys, attention-aware truncation cuts the
relative logit error $\lVert Q(K-K_r)^\top\rVert_F/\lVert QK^\top\rVert_F$ from $0.98$ to $0.27$ at
rank $r{=}4$, a $3.6\times$ reduction on the quantity attention actually consumes, even though its raw
$\lVert K-K_r\rVert_F$ is \emph{larger}. On the values, mass-weighted truncation lowers the output
error $\lVert P(V-V_r)\rVert_F/\lVert PV\rVert_F$ from $0.92$ to $0.64$ at $r{=}8$ and from $0.61$ to
$0.00$ at $r{=}16$, recovering the exact attention output once the rank spans the mass-weighted
subspace. The retained \emph{basis}, not just the rank, governs quality at a fixed memory budget.

\subsubsection{Real-world interpretability analysis}

\begin{figure}[p]
\centering
\includegraphics[width=0.80\linewidth]{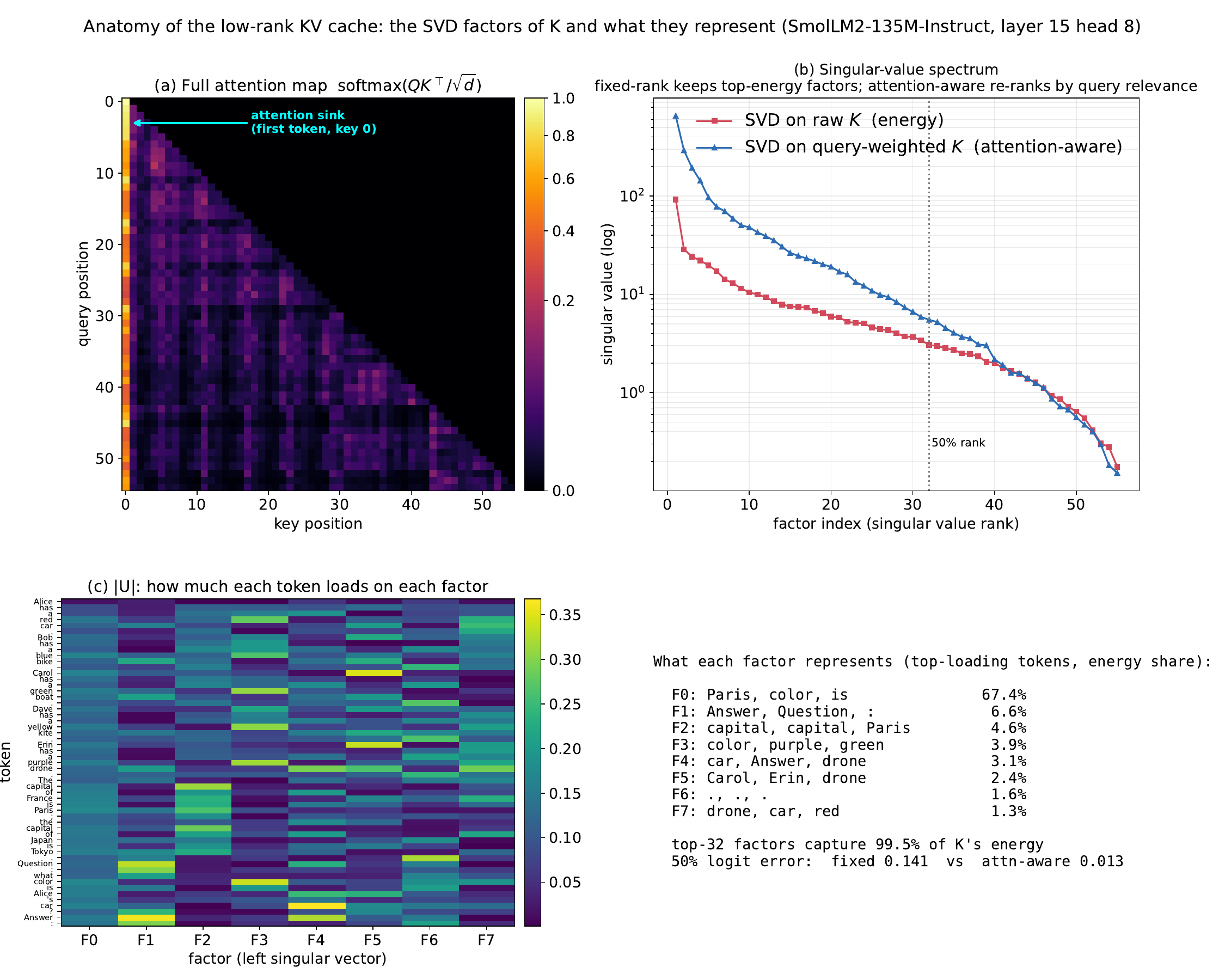}
\caption{SVD factors of the low-rank KV cache, $K$ (SmolLM2-135M, one layer/head).
\emph{(a)} attention map; \emph{(b)} $K$'s singular spectrum, SVD on raw KV (energy) vs.\ attention-aware
(query-weighted); \emph{(c)} per-token factor weights $|U|$; \emph{(d)} top tokens per factor with energy shares.}
\label{fig:svd-anatomy}

\vspace{0.5em}

\includegraphics[width=0.80\linewidth]{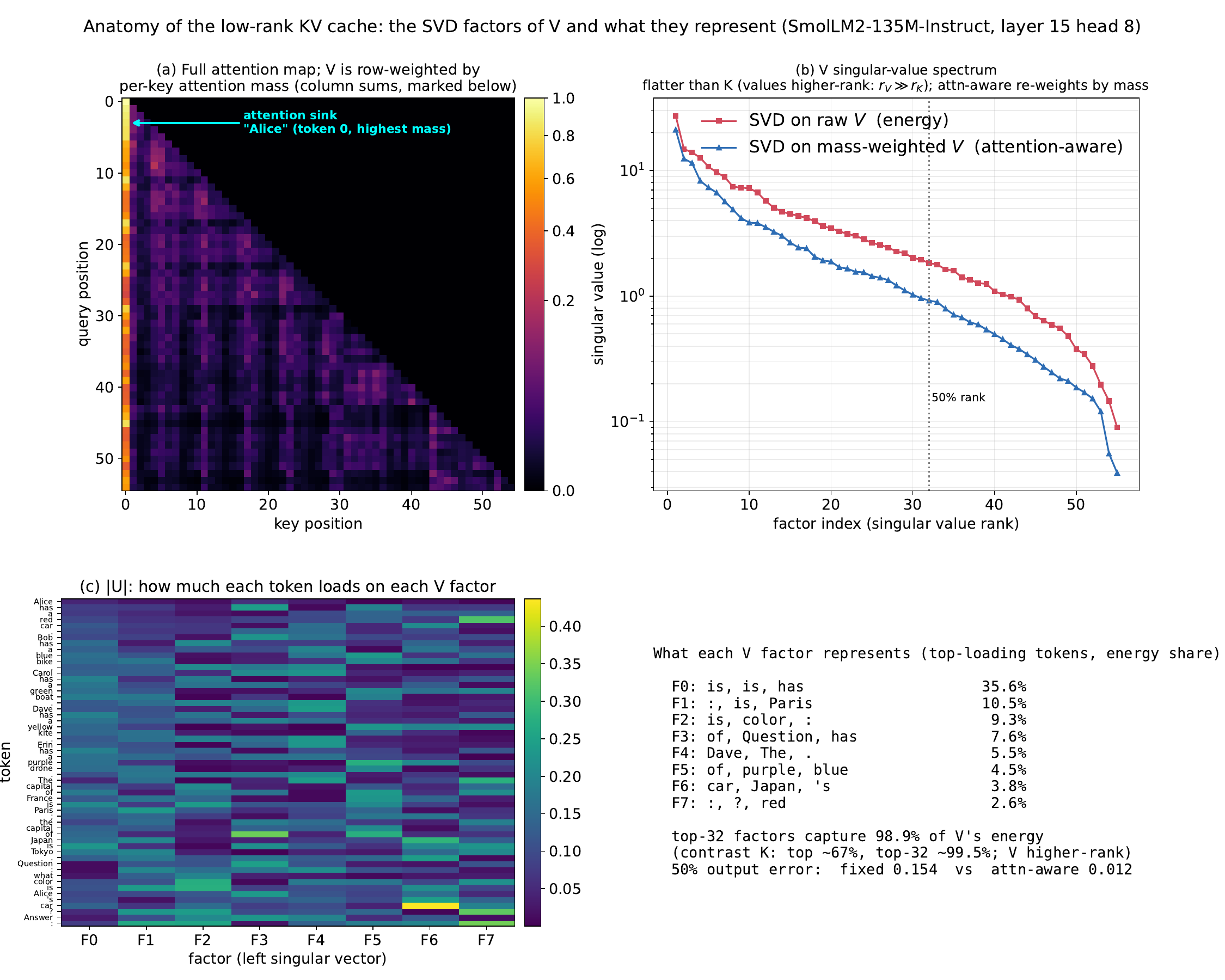}
\caption{SVD factors of the low-rank KV cache, $V$ (same layer/head and prompt as
Figure~\ref{fig:svd-anatomy}). \emph{(a)} attention map, with the per-key attention mass that weights $V$;
\emph{(b)} $V$'s singular spectrum, SVD on raw KV vs.\ attention-aware (mass-weighted); \emph{(c)} per-token
factor weights $|U|$; \emph{(d)} top tokens per factor with energy shares.}
\label{fig:svd-anatomy-v}
\end{figure}

\begin{figure}[H]
\centering
\includegraphics[width=\linewidth]{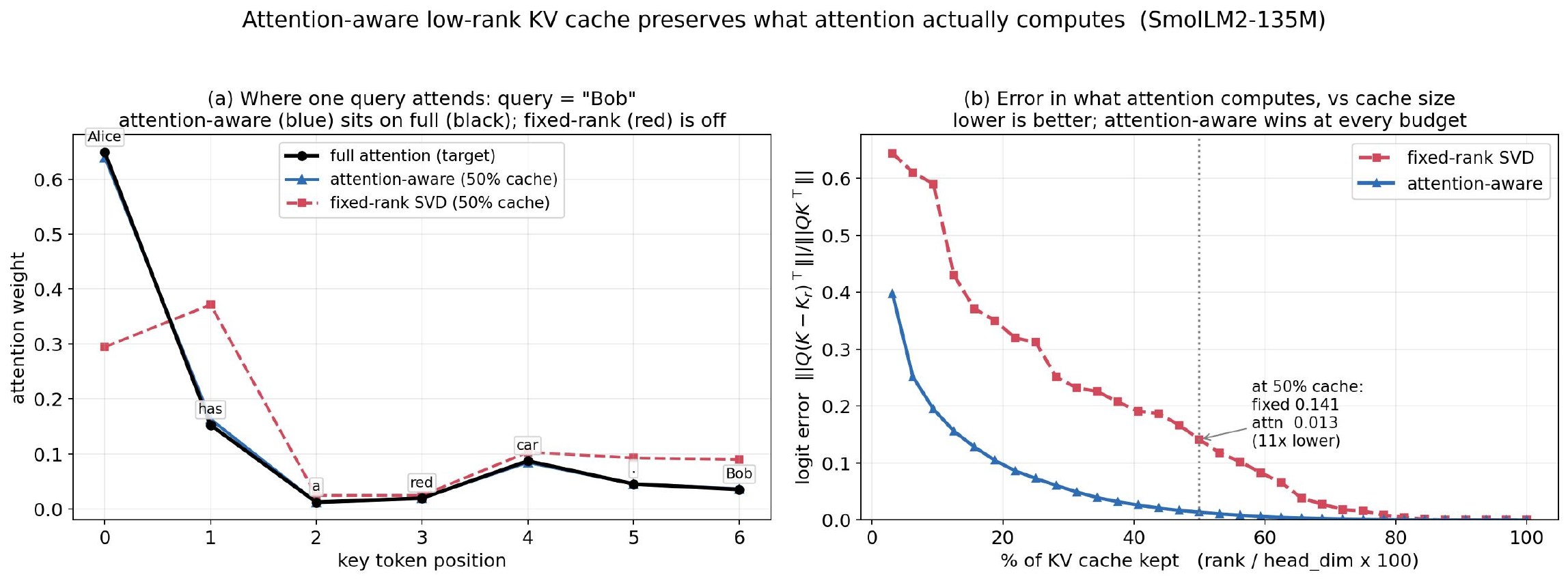}
\caption{The attention-aware cache preserves what attention computes (a real layer/head of
SmolLM2-135M). \emph{(a)} At $50\%$ cache the attention-aware keys (blue) lie almost on the
full-attention target (black), while SVD on raw KV (red) distorts it. \emph{(b)} Query-relevant logit
error vs.\ cache size: attention-aware stays below SVD on raw KV at every budget ($0.013$ vs.\ $0.141$ at
$50\%$). Real-model analogue of Table~\ref{tab:ablation}.}
\label{fig:attn-svd}
\end{figure}

We then move from the synthetic head to a real one. Figures~\ref{fig:attn-svd},
\ref{fig:svd-anatomy}, and~\ref{fig:svd-anatomy-v} inspect a single layer/head of
SmolLM2-135M\footnote{We use SmolLM2-135M for the factor-level and real-head
visualizations because its small hidden size and few heads keep a single head's attention map,
singular spectrum, and per-token factor weights legible in one figure, and the full per-head SVD runs on CPU;
the same attention-aware effect holds on the larger models used in our main quantitative
experiments.}
under both weightings. Figure~\ref{fig:attn-svd} contrasts the attention-aware and raw-KV caches
directly, while Figures~\ref{fig:svd-anatomy} and~\ref{fig:svd-anatomy-v} give a full
decomposition of $K$ and $V$ on that head; each reports four panels: \emph{(a)} the attention map,
\emph{(b)} the singular spectrum under raw vs.\ attention-aware weighting, \emph{(c)} the per-token
factor weights $|U|$, and \emph{(d)} the top-contributing tokens of each factor. The three quantities
of \S\ref{sec:interp} map onto this head as follows.

\paragraph{(i)~Effective rank.} Panel~(b) of each factor figure reports the per-matrix compression
capacity as the decay rate of its energy curve. The energy rule of Eq.~\eqref{eq:energyrank} measures
this decay on the attention-aware matrices $\widetilde K$ and $\widetilde V$, not the raw $K$ and
$V$, and sets each its own rank accordingly. $\widetilde K$'s spectrum is steep, with ${\sim}67\%$ of
the energy in a single leading factor and $99.5\%$ in the top-$32$ (a $50\%$ cache), whereas
$\widetilde V$'s spectrum is markedly flatter, with only ${\sim}36\%$ in its leading factor. The raw
$K$ and $V$ spectra (red in panel~(b)) track these attention-aware curves closely in overall decay,
so the steep versus flat gap is intrinsic to this head rather than an artifact of the weighting; the
query-weighted $\widetilde K$ curve is only somewhat sharper, concentrating on the query-relevant
directions that item~(ii) examines. That asymmetry is
the concrete per-head cause of the $r_{\widetilde K}\!\ll\!r_{\widetilde V}$ allocation that
Eq.~\eqref{eq:energyrank} exploits, with $\widetilde V$ genuinely less compressible than
$\widetilde K$.

\paragraph{(ii)~Query subspace.} Panel~(b) also shows that the attention-aware basis re-ranks factors
by query relevance rather than raw energy, keeping a \emph{different} rank-$r$ subspace than SVD on raw KV; the gap between the raw-energy (red) and attention-weighted (blue) curves makes visible that
ordering factors by energy is not the same as ordering them by what the queries consume.
Figure~\ref{fig:attn-svd} confirms on the real head the same effect Table~\ref{tab:ablation} showed on
the synthetic one: at $50\%$ cache the attention-aware keys (blue, panel~(a)) sit almost exactly on
the full-attention distribution (black) while SVD on raw KV (red) distorts it, and the query-relevant
logit error stays more than $10\times$ lower across budgets ($0.013$ vs.\ $0.141$ at $50\%$,
panel~(b)).

\paragraph{(iii)~Token saliency.} Panels~(c) and~(d) tie the factors back to tokens. The per-token
factor weights $|U|$ in panel~(c) group the leading $K$ factors into roughly semantic clusters, for example
a colour factor $\{$color, purple, green$\}$ identified from these weights, and panel~(d) lists the
top-contributing tokens of each factor with their energy shares. Row-weighting each token by its attention
mass $a_i{=}\sum_q P_{qi}$ reconstructs the heavily attended
tokens, such as the first-token sink, most faithfully, lowering
$\lVert P(V-V_r)\rVert_F/\lVert PV\rVert_F$ from $0.154$ to $0.012$ at $50\%$ cache, roughly
$13\times$. End-to-end on Phi-4-14B agentic benchmark (Figure~\ref{fig:phi4sweep}) the same attention-aware
basis adds $+0.13$ to $0.29$ mean judge and $+7$ to $18$ points $\mathrm{pass@}4$ over SVD on raw KV,
reaching $4.067$ / $62.5\%$ at $50\%$ cache, within $0.08$ of dense.

Taken together, the three items justify all of \method's design choices at once: low-rank caching
(both spectra decay), asymmetric per-matrix ranks ($r_{\widetilde K}\!\ll\!r_{\widetilde V}$), and the attention-aware
basis (blue $\ne$ red).

\end{document}